\documentclass[10pt,twocolumn,letterpaper]{article}

\usepackage{cvpr}              

\usepackage{multirow}
\usepackage{threeparttable}
\definecolor{cvprblue}{rgb}{0.21,0.49,0.74}
\usepackage[pagebackref,breaklinks,colorlinks,allcolors=cvprblue]
{hyperref}

\usepackage{multirow}
\usepackage{tikz}
\usepackage{makecell}
\usepackage{mathtools}
\usepackage{algorithm}
\usepackage{algpseudocode}
\usepackage{amsmath}

\def\paperID{} 
\def\confName{CVPR}
\def\confYear{2026}

\newcommand{\bK}{\mathbf{K}}

\newcommand{\bx}{\mathbf{x}}

\newcommand{\bI}{\mathbf{I}}
\newcommand{\bzero}{\mathbf{0}}

\newcommand{\bV}{\mathbf{V}}

\newcommand{\bv}{\mathbf{v}}

\newcommand{\bR}{\mathbf{R}}

\newcommand{\bz}{\mathbf{z}}
\newcommand{\bP}{\mathbf{P}}
\newcommand{\bC}{\mathbf{C}}

\newcommand{\bc}{\mathbf{c}}

\newcommand{\bT}{\mathbf{T}}

\newcommand{\beps}{\boldsymbol{\epsilon}}

\newcommand{\nR}{\mathbb{R}}

\newcommand{\E}{\mathbb{E}}
\newcommand{\cN}{\mathcal{N}}

\newcommand{\cL}{\mathcal{L}}

\newcommand{\cE}{\mathcal{E}}
\newcommand{\cD}{\mathcal{D}}

\makeatletter
\DeclareRobustCommand\onedot{\futurelet\@let@token\@onedot}
\def\@onedot{\ifx\@let@token.\else.\null\fi\xspace}
\def\Fig{Fig\onedot}   
\makeatother

\newcommand{\figref}[1]{\Fig~\ref{#1}}
\newcommand{\secref}[1]{Sec.~\ref{#1}}

\newcommand{\algsref}[1]{Algorithm~\ref{#1}}
\renewcommand{\eqref}[1]{Eq.~\ref{#1}}
\newcommand{\tabref}[1]{Tab.~\ref{#1}}

\newif\ifcomment
\commenttrue
\ifcomment
	
\fi

\title{Plenoptic Video Generation}

\author{Xiao Fu$^{1,2}$, 
Shitao Tang$^{1}$,
Min Shi$^{1,3}$,
Xian Liu$^{1}$,
Jinwei Gu$^{1}$, 
Ming-Yu Liu$^1$,
Dahua Lin$^2$, 
Chen-Hsuan Lin$^1$\\
$^1$NVIDIA,
$^2$The Chinese University of Hong Kong,
$^3$Georgia Institute of Technology \\
}

\begin{document}

\twocolumn[{
\renewcommand\twocolumn[1][]{#1}
    \maketitle
    \vspace{-2em}
    \begin{center}
        \includegraphics[width=\linewidth]{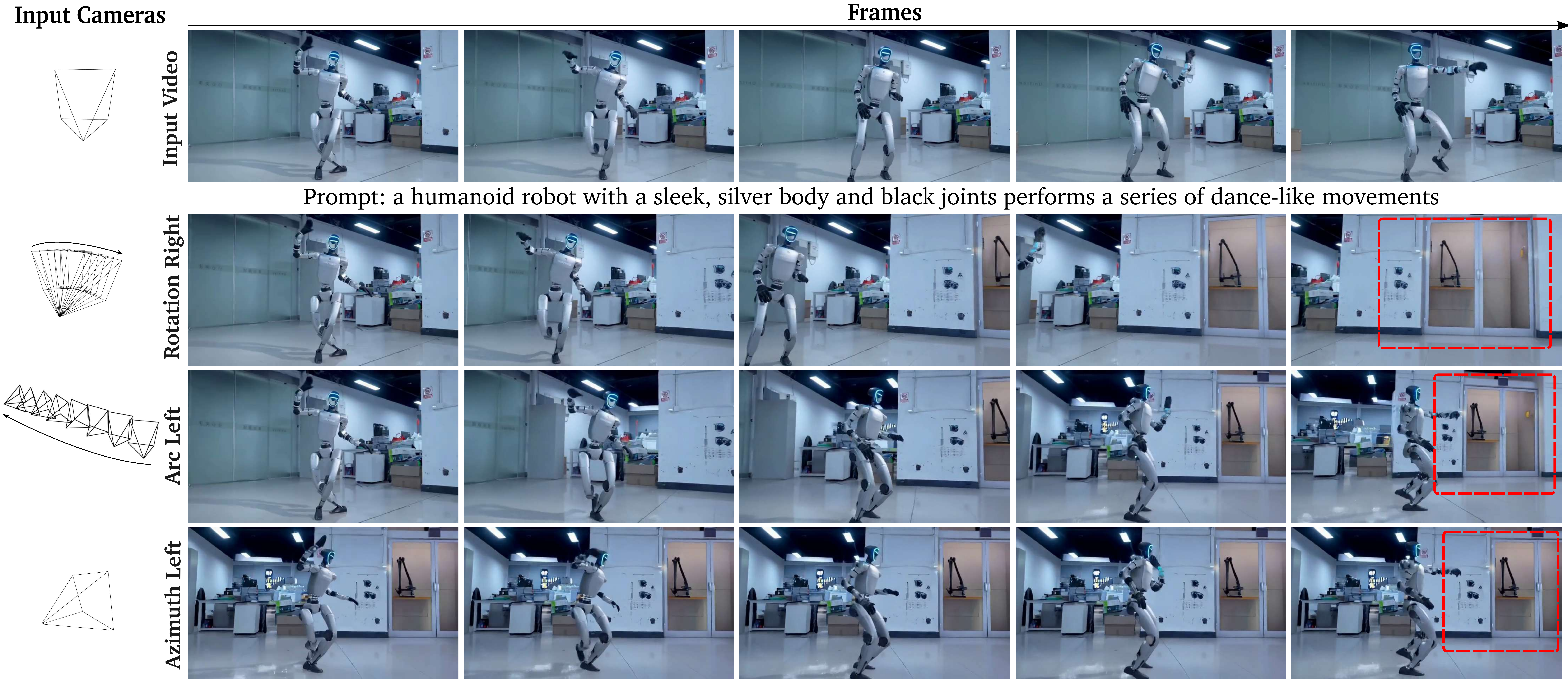}
        \setlength{\abovecaptionskip}{0mm}
        \captionof{figure}{\small
        We present \textit{PlenopticDreamer}, a generative framework that re-renders input video under novel camera trajectories while preserving long-term spatio-temporal memory in hallucinated regions across overlapping views, thereby producing coherent plenoptic functions (see robot’s right side, highlighted in red dashed boxes across three trajectories). Please refer to our~\href{https://research.nvidia.com/labs/dir/plenopticdreamer/}{website} for more results.        
        }
        \label{fig:teaser}
    \end{center}
}]

\begin{abstract}

Camera-controlled generative video re-rendering methods, such as ReCamMaster, have achieved remarkable progress. However, despite their success in single-view setting, these works often struggle to maintain consistency across multi-view scenarios. Ensuring spatio-temporal coherence in hallucinated regions remains challenging due to the inherent stochasticity of generative models. To address it, we introduce PlenopticDreamer, a framework that synchronizes generative hallucinations to maintain spatio-temporal memory. The core idea is to train a multi-in-single-out video-conditioned model in an autoregressive manner, aided by a camera-guided video retrieval strategy that adaptively selects salient videos from previous generations as conditional inputs. In addition, Our training incorporates progressive context-scaling to improve convergence, self-conditioning to enhance robustness against long-range visual degradation caused by error accumulation, and a long-video conditioning mechanism to support extended video generation. Extensive experiments on the Basic and Agibot benchmarks demonstrate that PlenopticDreamer achieves state-of-the-art video re-rendering, delivering superior view synchronization, high-fidelity visuals, accurate camera control, and diverse view transformations (\eg, third-person → third-person, and head-view → gripper-view in robotic manipulation). Project page:~\url{https://research.nvidia.com/labs/dir/plenopticdreamer/}.

\end{abstract}

\section{Introduction}

Video generation~\cite{guo2023animatediff,ali2025world,wan2025wan,kong2024hunyuanvideo,hong2022cogvideo,brooks2024video,peebles2023scalable,kling2025videoturbo,gao2025seedance,veo2025google,fu2025learning} has become increasingly prevalent in content creation and social media.  As video frames result from camera projections of scene radiance, they can be interpreted as discrete samples of the underlying plenoptic function~\cite{bergen1991plenoptic,li2020crowdsampling}. Consequently, effective control of camera motion~\cite{wang2024motionctrl,he2025cameractrl,bahmani2025ac3d,ren2025gen3c} is essential for shaping the captured light field, emphasizing visual focus, and guiding the viewer’s attention.

Recently, camera-controlled generative video re-rendering, which aims to synthesize novel videos along arbitrary camera trajectories while preserving the original content, has attracted significant attention, supporting applications such as immersive content creation and embodied AI. Representative methods, including ReCamMaster~\cite{bai2025recammaster} and TrajectoryCrafter~\cite{yu2025trajectorycrafter}, achieve promising results on their curated real-world or synthetic datasets. However, these methods primarily succeed in the~\textit{single-view} setting and struggle in~\textit{multi-view} scenarios, which are essential for reconstructing a holistic representation of the scene.
Specifically, they fail to maintain consistent spatio-temporal hallucinations in regions unseen from the source view. The inherent stochasticity of diffusion models, combined with their limited long-range spatial memory, leads to geometric misalignment and view desynchronization across different camera-conditioned generations.

To address it, we present~\textit{PlenopticDreamer}, a camera-controlled generative video re-rendering framework that explicitly enforces spatio-temporal memory for consistent scene generation. Unlike prior single-shot methods that generate each view independently, PlenopticDreamer adopts an~\textit{autoregressive}, multi-in–single-out formulation. At each step, it retrieves a set of previously generated video–camera pairs from a memory bank and conditions the next generation on these retrieved contexts. This design enables synchronized hallucinations across time and viewpoints while preserving scene geometry and motion dynamics. Video context retrieval is guided by a 3D field-of-view (FOV) mechanism that evaluates spatial co-visibility to select the most relevant past video segments.

In addition, PlenopticDreamer introduces two training strategies that substantially improve robustness and convergence.
First, progressive context-scaling stabilizes optimization by gradually increasing the number of conditioning videos during training, enabling the model to learn context-aware reasoning across short to long temporal horizons. Second, self-conditioned training mitigates error accumulation in autoregressive generation by fine-tuning the model on its own synthesized outputs. Together, these strategies facilitate stable video synthesis while preserving spatial alignment and temporal consistency.
We further propose a long-video conditioning mechanism to extend the model’s capability to render longer video sequences.

We evaluate our method on two benchmarks: (1) a~\textit{Basic benchmark} covering diverse in-the-wild scenes, and (2) an~\textit{Agibot benchmark}~\cite{bu2025agibot} focusing on robotic manipulation.
Experimental results show that PlenopticDreamer achieves state-of-the-art performance in view synchronization while maintaining accurate camera control and high-fidelity visual quality. In summary, our contributions are:
\begin{enumerate}[itemsep=0em]
    \item We present PlenopticDreamer, the first camera-controlled generative video re-rendering framework with long-term spatio-temporal memory.
    \item We propose an autoregressive architecture with a 3D FOV–based video retrieval mechanism for scalable, coherent multi-camera generation. We incorporate progressive context-scaling and self-conditioned training strategies to enhance stability and long-term consistency.
    \item Extensive experiments show that our method achieves state-of-the-art performance in video re-rendering, including view synchronization, camera accuracy, and visual fidelity.
    It supports diverse camera transformations (\eg, third-person → third-person, head-view → gripper-view) and enables long video generation.
\end{enumerate}

\section{Related Work}

\noindent\textbf{Camera-Controlled Video Generation.}
Effective control of camera motion has received significant attention in video generation. Existing approaches can be broadly categorized into three directions:
(1)~\textit{single-view generation}: works rely on explicit 6DoF camera poses~\cite{wang2024motionctrl,ren2025gen3c} or pixel-wise Plücker raymaps~\cite{he2025cameractrl,bahmani2025ac3d,bahmani2024vd3d,zheng2024cami2v,xu2024camco} to guide text- or image-to-video synthesis, achieving view control. Methods~\cite{hu2024motionmaster,hou2024training,ling2024motionclone,yang2024direct} explore training-free strategies for camera manipulation, and a few~\cite{fu20243dtrajmaster,wang2024objctrl} extend these mechanisms to control object motion via camera trajectories.
(2)~\textit{multi-view video generation}: methods aim to maintain cross-view consistency, ranging from object-level synthesis~\cite{li2024vivid,xie2024sv4d} to scene-level reconstruction~\cite{kuang2024collaborative,bai2024syncammaster}.
(3)~\textit{video-to-video re-rendering}: some approaches~\cite{van2024generative,bai2025recammaster} perform implicit re-rendering with minimal 3D supervision, whereas others~\cite{yu2025trajectorycrafter,xiao2024trajectory,xie2024sv4d,gu2025diffusion,ren2025gen3c} project video context into 3D representations to synthesize novel views. Despite these advances, none integrate memory mechanisms to maintain long-term spatio-temporal coherence across multiple views. In contrast, our PlenopticDreamer introduces the first memory-based framework for generative video-to-video re-rendering, achieving coherent multiview synthesis.
\begin{figure*}[!tp]
\centering
\includegraphics[width=0.99\textwidth]{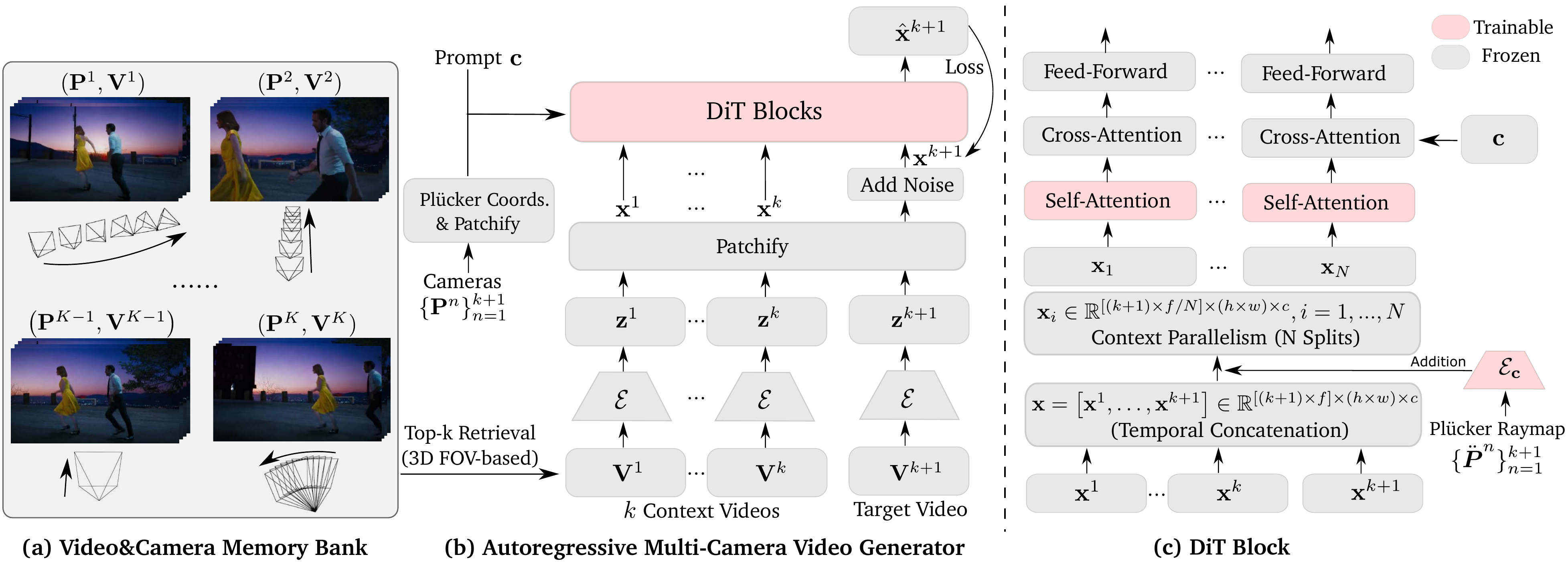}
\caption{\textbf{PlenopticDreamer Framework.} Its core is an autoregressive multi-camera video generator that retrieves $k$ video–camera pairs $\{(\bP^n, \bV^n)\}_{n=1}^k$ from the memory bank using a 3D FOV–based retrieval strategy. Conditioned on these retrieved pairs and the target camera $\bP^{k+1}$, the model performs noisy scheduling and learnable reconstruction to generate the target video $\bV^{k+1}$. To enable long video generation, a portion of the preceding frames in $\bV^{k+1}$ is preserved as clean inputs at a certain ratio during training. Within each DiT block, temporal concatenation is applied to form video tokens $\bx$ as in-context condition.}

\label{fig:pipeline}
\end{figure*}

\vspace{-1em}
\noindent\textbf{Memory Mechanism for Video Generation.}
Building long-term memory is essential for coherent video generation~\cite{song2025history,kanervisto2025world}.
Existing approaches can be broadly categorized into four types:
(1)~\textit{frame-level memory}: methods such as~\cite{xiao2025worldmem,chen2025learning,yu2025context,schneider2025worldexplorer,chen2025deepverse} store key historical frames and retrieve the top-$k$ relevant ones via camera-pose similarity for conditioning;
(2)~\textit{latent-level memory}: approaches including~\cite{liu2025worldweaver,zhang2025packing,li2025hunyuan,mao2025yume,cai2025mixture} maintain hierarchical memory capturing long-term coarse tokens and short-term fine-grained tokens, adaptively retrieving salient features during inference;
(3)~\textit{3D-level memory}: works like VMem~\cite{li2025vmem} and SPMem~\cite{wu2025video} reconstruct 3D structures (\eg, surfels or point clouds) to store video context and render geometry-aware representations for novel-view synthesis;
(4)~\textit{network-level memory:} TTT-Video~\cite{dalal2025one} leverages Test-Time Training (TTT) layers to record input tokens and update model weights.
In contrast, our PlenopticDreamer introduces aN explicit \textit{video-based retrieval} mechanism conditioning generation on camera-guided selection of past video segments.

\section{Method}

Our goal is to enable generative video-to-video re-rendering with spatio-temporal memory, interpretable as generating the space-time dependent plenoptic function of a scene. We first introduce the preliminaries and task formulation in~\secref{sec:preliminary}, followed by our autoregressive modeling paradigm and video retrieval mechanism in~\secref{sec:inject_condition}. Finally, we describe the enhanced training strategies in~\secref{sec:train_strategy}. The overall framework is illustrated in~\figref{fig:pipeline}.

\subsection{Preliminary and Problem Definition} \label{sec:preliminary}

\noindent\textbf{Flow-based Video Diffusion Transformer (DiT).}
We conduct experiments using a video diffusion transformer model under the flow-matching paradigm~\cite{lipman2022flow,esser2024scaling}.
Given a data sample $\bx_{0} \sim p(\bx)$, a noise sampler $\beps \sim \cN\left(\bzero, \bI\right)$, and a continuous time variable $t \in[0,1]$, the forward process linearly interpolates between data and noise distribution, \ie,
\begin{equation}
\bx_t=(1-t) \bx_0+t \beps, \bv_t=\beps-\bx_0
\end{equation}
where $\bv_t$ is the GT velocity field.
The denoising process is solved by an ordinary differential equation (ODE):
\begin{equation}
d \bx_t=\bv_\Theta\left(\bx_t, t, \bc\right) d t
\end{equation}
where $\bv_\Theta (\cdot)$ represents the predicted velocity function parameterized by a transformer-based network~\citep{peebles2023scalable}, and $\bc$ is the conditional signal (\eg, video context).
The model is optimized using the following flow-matching objective:
\begin{equation}
\cL(\Theta)=\E_{\bx, \beps, \bc, t}\left\|\bv_\Theta\left(\bx_t, t, \bc \right)-\bv_t\right\|^2
\end{equation}
During training, the timestep $t$ can be biased toward higher noise levels to encourage robust reconstruction under degraded spatio-temporal correlations.

\noindent\textbf{Task Formulation and Notations.}
Given a source video $\bV_s \in \nR^{F \times C \times H \times W}$ and a set of $N$ target camera trajectories $\{\bP_t^n\}_{n=1}^N$, each $\bP_t$ specified by extrinsic parameters $\bC_t=[\bR_t; \bT_t] \in \nR^{F \times 3 \times 4}$ and intrinsics $\bK_t \in \nR^{3 \times 3}$, our objective is to synthesize $N$ target videos $\{\bV_t^n\}_{n=1}^N$, with each $\bV_t^n \in \nR^{F \times C \times H \times W}$ sharing the same context as the input video while corresponding to a distinct virtual camera trajectory.
The generated videos are required to maintain the source content fidelity and exhibit synchronized spatial-temporal consistency across viewpoints, particularly in hallucinated regions.
We employ a standard pinhole camera model (zero horizontal and vertical skew) for generation.
The overall generative process $f(\cdot)$ is formulated as
\begin{equation} \label{eq:task}
f(\cdot): \bc, \bV_s, \bP_s, \{\bP_t^n\}_{n=1}^N \rightarrow \{\bV_t^n\}_{n=1}^N
\end{equation}
where $\bc$ denotes the video caption and $\bP_s$ is the camera trajectory of the source video.
 Additionally, a variational autoencoder with encoder $\cE(\cdot)$ and decoder $\cD(\cdot)$ is employed to map the videos between pixel-space and latent-space ($\bV_s \leftrightarrow \bz_s, \{\bV_t^n\}_{n=1}^N \leftrightarrow \{\bz_t^n\}_{n=1}^N$, where $\bz \in \nR^{f \times h \times w \times c}$).

\subsection{Injecting Conditions into Video DiT} \label{sec:inject_condition}

To effectively guide the video model with target conditions, including the source video and target camera trajectories, we adopt an in-context conditioning strategy.~\cite{ju2025fulldit,bai2025recammaster,yu2025context,he2025fulldit2}.

\noindent\textbf{Native Solution.}
A straightforward approach is to enlarge the context window by extending the number of input videos from 1 to $N$, following ReCamMaster~\cite{bai2025recammaster}:
\begin{equation}
\left\{\begin{aligned}
\bx_s & =\operatorname{patchify}\left(\bz_s\right), \bx_t^n=\operatorname{patchify}\left(\bz_t^n\right), n=1,...,N \\
\bx& =\left[\bx_s, \bx_1, ..., \bx_N\right]_{\text {frame-dim }} \in \nR^{[(N+1) \times f] \times (h \times w) \times c},
\end{aligned}\right.
\end{equation}
where $\bx$ denotes the input video tokens fed into the DiT block.
While this strategy can be effective for small $N$ (\eg, 2 or 3) under low-resolution settings ($\leq$480p), it rapidly becomes computationally prohibitive and prone to out-of-memory (OOM) failures as $N$ or video resolution increases.

\noindent\textbf{Autoregressive Generation Paradigm.} 
Inspired by the effectiveness of the autoregressive paradigm for long-context modeling~\cite{yu2025context,wang2024loong,achiam2023gpt,chen2024diffusion,huang2025self}, we reformulate video generation as a sequential process instead of performing single-shot inference. Specifically, we generate one video at a time and produce all videos in a sequential manner. 
Accordingly, we rewrite~\eqref{eq:task} as:
\begin{equation} \label{eq:task_new_1}
f(\cdot): \bc, \{\left(\bP^n, \bV^n\right)\}_{n=1}^k, \bP^\text{k+1} \rightarrow \bV^\text{k+1}, k=1,...,N-1
\end{equation}
where $\{(\bP^n, \bV^n)\}_{n=1}^k$ denotes previously generated videos and their cameras and $(\bP_s, \bV_s)$ is regarded as $(\bP^1, \bV^1)$, and $k$ is the model context size.
We adopt temporal concatenation strategy to form conditional video tokens:
\begin{equation}
\bx =\left[\bx_1, ..., \bx_\text{k+1}\right]_{\text {frame-dim}} \in \nR^{[(k+1) \times f] \times (h \times w) \times c}
\end{equation}

\noindent\textbf{Camera Conditioning.} 
To encode camera information, we represent it using Plücker raymaps~\citep{sitzmann2021light}, mapping pixels to 6D ray representations: $\bP^n=(\bC^{n}\in \nR^{f \times 3 \times 4}, \bK \in \nR^{3 \times 3}) \rightarrow \ddot{\bP}^n \in \nR^{f \times H \times W \times 6}, n=1,...,k+1$.
These raymaps are then temporally concatenated and patchified. A camera projection layer $\cE_{\text{cam}}(\cdot)$ is introduced to align the raymap dimensionality with that of the video latents. The resulting raymap tokens are channel-wise added to the video tokens before self-attention layer, enabling DiT to integrate camera pose information.

\begin{figure}[!th]
\centering
\includegraphics[width=0.49\textwidth]{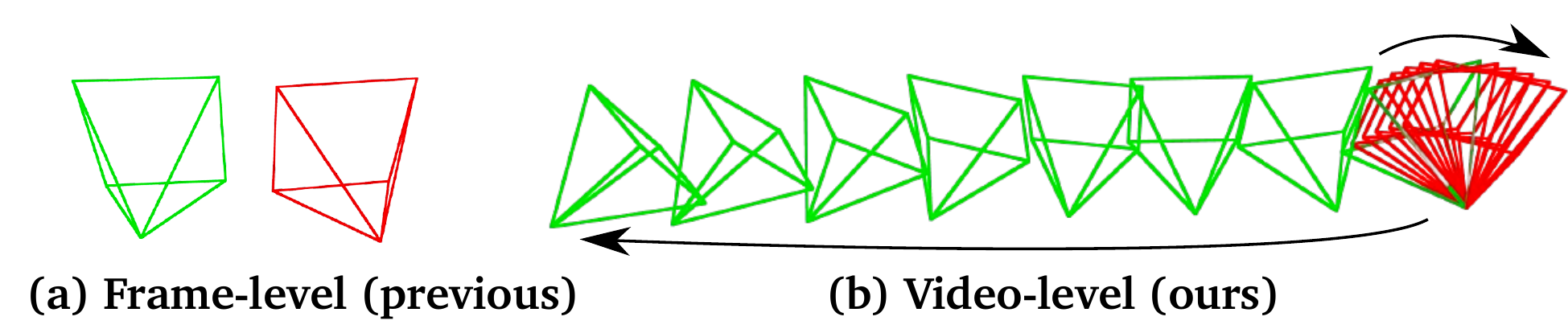}
\caption{\textbf{FOV-based Retrieval Comparison.} Unlike prior frame-level retrieval methods~\cite{yu2025context,xiao2025worldmem}, ours computes robust video-level similarity by averaging frame-wise similarities.}
\label{fig:video_fov}
\end{figure}

\noindent\textbf{3D FOV–based Video Retrieval.}
A key challenge in this autoregressive framework lies in selecting the most salient $k$ videos from the previously video pool as conditioning inputs for the next generation step. Given that each video corresponds to a distinct camera trajectory, we adopt a 3D Field-of-View (FOV) retrieval mechanism to identify the most relevant candidates. Specifically, we compute video-level similarity via spatial co-visibility across all frames, and select the top-$k$ context videos, as illustrated in~\algsref{alg:memory_retrieval} and~\figref{fig:video_fov}. When the number of context videos is less than $k$, we replicate input video-camera pair $(\bP^1, \bV^1)$ to match the required context length.
Furthermore, when the number of retrieved videos $l$ exceeds the model’s context capacity $k$, a divide-and-conquer inference strategy is employed to cover as diverse viewpoints as possible and minimize viewpoint overlap, as described in~\algsref{alg:divide_conquer}. Here the trajectory fusion process results in a merged trajectory roughly spanning the FOV of all inputs.

\begin{algorithm}[t]
\caption{Video Retrieval Algorithm}
\label{alg:memory_retrieval}
\begin{algorithmic}[1]
\Statex \hspace*{-\algorithmicindent}\textbf{Input:}
\begin{itemize}
  \item Memory bank of $K$ videos $\{(\bV^{n}, \bP^{n})\}_{n=1}^K$
  \item Target camera trajectory $\bP^{\text{K+1}}$
  \item Maximum retrieved video number $k$
  \item Near/Far plane distances $D_n, D_f$
  \item Monte Carlo sampling points $P$
\end{itemize}
\Statex \hspace*{-\algorithmicindent}\textbf{Output:} Top-$k$ retrieved videos
\State Initialize similarity set $S \gets \varnothing$
\For{$n = 1$ to $K$}
\State Initialize similarity $S_n \gets 0$
\For{$f = 1$ to $F$}
    \State Construct $f$-th camera frustum of $\bP^{n}$ and $\bP^{\text{K+1}}$
    \State Perform Monte Carlo sampling within near/far planes of each frustum
    \State Count visible points $P_n$, $P_{\text{K+1}}$ in other's frustum
    \State Update similarity: $S_n \gets S_n + \frac{P_n + P_{K+1}}{2P \times F}$
\EndFor
\State Append $S_n$ to $S$
\EndFor
\State Select indices of the top-$k$ values in $S$
\State \Return retrieved $k$ videos
\end{algorithmic}
\end{algorithm}

\begin{algorithm}[t]
\caption{Divide-and-Conquer Inference Algorithm}
\label{alg:divide_conquer}
\begin{algorithmic}[1]
\Statex \hspace*{-\algorithmicindent}\textbf{Input:}
\begin{itemize}
  \item Top-$l$ retrieved videos $\{(\bV^{n}, \bP^{n})\}_{n=1}^{l}$ (sorted by ascending camera similarity)
  \item Model context video size $k$
  \item Target camera trajectory $\bP^{\text{L+1}}$
\end{itemize}
\Statex \hspace*{-\algorithmicindent}\textbf{Output:} Target video $\bV^{\text{L+1}}$.
\While{$l > k$}
    \State Select the first $m = \min(l-k, k)$ videos to form context set $V$
    \While{$m < k$}
        \State Append $(\bP_s, \bV_s)$ to $V$
        \State $m \gets m + 1$
    \EndWhile
    \State Merge trajectories in $V$ to form $\bP_{\text{merge}}$
    \State Infer merged video $\bV_{\text{merge}}$ using $(V, \bP_{\text{merge}})$
    \State Replace the first $m$ elements with $(\bV_{\text{merge}}, \bP_{\text{merge}})$
    \State $l \gets l - m + 1$
\EndWhile
\State Perform final inference to obtain $\bV^{L+1}$
\State \Return target video $\bV^{L+1}$
\end{algorithmic}
\end{algorithm}

\noindent\textbf{Autoregressive Long Video Generation.}
For input video exceeding the model’s temporal window, we partition them into overlapping sub-chunks $\{\bV_s^{m}\}_{m=1}^M$, where consecutive chunks share a set of frames from the latter portion of the previous chunk to preserve temporal continuity.
Unlike the formulation in~\eqref{eq:task_new_1}, where $\bV^\text{k+1}$ is generated from pure noise, we incorporate the overlapping frames as additional conditioning:
\vspace{-0.5em}
\begin{equation} \label{eq:task_new_2}
f(\cdot): \bc, \{(\bP^{n,m}, \bV^{n,m})\}_{n=1}^k, \bP^\text{k+1,m}, \tilde{\bV}^\text{k+1,m} \rightarrow \bV^\text{k+1,m}
\end{equation}
where $k=1,...,N-1, m=1,...,M$. Here $\tilde{\bV}^{k+1,m} \in \nR^{\tilde{F} \times C \times H \times W}$ contains the $\tilde{F}$ overlapping frames, and $m$ indexes the $m$-th sub-chunk $\bV_s^{m}$ from the source video.
During inference, we sequentially generate the videos as follow:
\begin{equation}
\underbrace{\bV^\text{1,1} \rightarrow  \bV^\text{2,1} \rightarrow ...
\rightarrow \bV^\text{N,1}}_{\text{Finish the first chunk $\bV_s^{1}$}} \underbrace{\rightarrow \bV^\text{1,2} \rightarrow \bV^\text{2,2} ...}_{\text{Start the next chunk $\bV_s^{2}$}} \rightarrow \bV^\text{N,M}
\end{equation}

\subsection{Training Strategy} \label{sec:train_strategy}

\noindent\textbf{Progressive Training.}
The training objective corresponding to~\eqref{eq:task_new_1} is defined as:
\begin{equation} \label{eq:loss}
\cL(\Theta)=\E_{\beps, \bc, \bP, \bV, t}\left\|\bv_\Theta\left(\{(\bP^n, \bV^n)\}_{n=1}^{\text{k+1}}, t, \bc \right)-\bv_t\right\|^2
\end{equation}
We also incorporate the extended prediction $\tilde{\bV}^{k+1}$ from~\eqref{eq:task_new_2} into the loss at a certain ratio.
In our empirical experiments, we observe that directly training the model with a large context size often leads to unstable convergence.
To address this, we adopt a progressive training strategy: the model is first trained with a small context size (\eg, $1$) and gradually scaled up as training stabilizes, until reaching the target context size $k$.
This progressive scheme significantly improves convergence stability and accelerates training in later stages with larger contexts.

\begin{table*}[!th] 
\caption{\textbf{Quantitative Comparison on the Basic Benchmark.} Ours consistently outperforms all baselines in view synchronization across all shots while maintaining high-fidelity visual quality and accurate camera accuracy. ReCamMaster* denotes a retrained version on Cosmos-Predict2.5 with Plücker raymaps, using the same combined datasets (MultiCamVideo + SyncCamVideo) for a fair comparison.} \label{tab:main_comparison} 
\centering 
\begin{threeparttable}
\resizebox{.89\textwidth}{!} 
{ \begin{tabular}{ccccccccc} 
\toprule 
& \multicolumn{1}{c}{ Visual Quality } & \multicolumn{2}{c}{ Camera Accuracy } & \multicolumn{4}{c}{ View Synchronization (Mat. Pix.(K) $\uparrow$)} \\ \cmidrule(lr){2-2} \cmidrule(lr){3-4} \cmidrule(lr){5-8} Model & FVD $\downarrow$ & TransErr $\downarrow$ & RotErr (rad) $\downarrow$ & 3 Shots & 6 Shots & 9 Shots & 12 Shots \\ 
\midrule 
Trajectory-Attention~\cite{xiao2024trajectory} & 734.1 & 0.77 & 0.26 & 22.7 & 26.9 & 28.8 & 29.1 \\
TrajectoryCrafter~\cite{yu2025trajectorycrafter} & \underline{665.9} & 0.65 & 0.27 & 31.2 & \underline{29.3} & \underline{35.3} & \underline{36.2} \\ 
ReCamMaster~\cite{bai2025recammaster} & 731.6 & 0.72 & 0.23 & \underline{32.1} & 29.0 & 30.9 & 27.6 \\ 
ReCamMaster*~\cite{bai2025recammaster} & 675.4 & \textbf{0.52} & \underline{0.22} & 24.6 & 20.2 & 29.7 & 31.2 \\ 
\midrule
PlenopticDreamer (Ours) & \textbf{425.8} & \underline{0.54} & \textbf{0.21} & \textbf{41.4} & \textbf{40.8} & \textbf{45.4} & \textbf{41.2} \\ 
\bottomrule 
\end{tabular} } 
\end{threeparttable} 
\end{table*}
\begin{figure*}[!th]
\centering
\includegraphics[width=\textwidth]{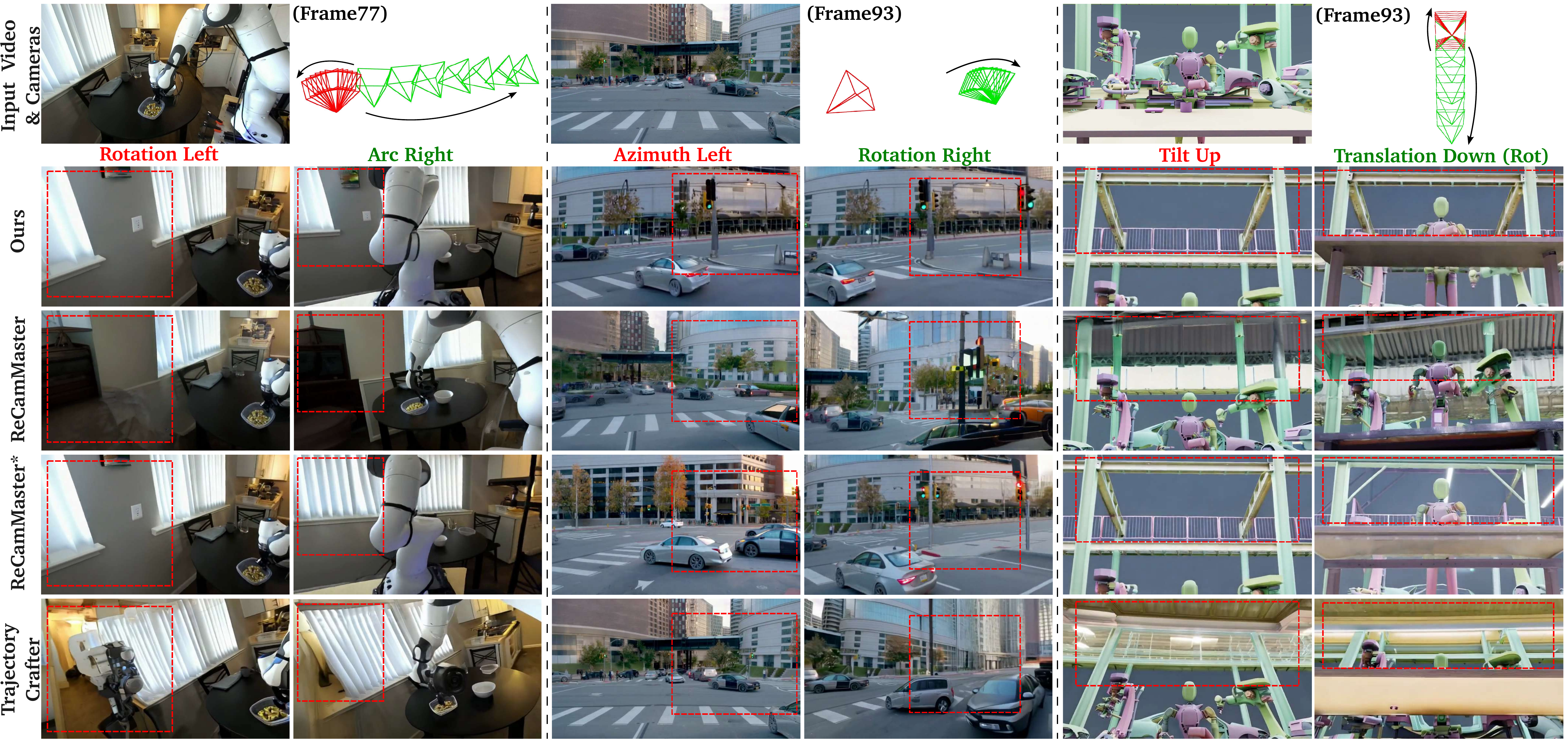}
\caption{\textbf{Qualitative Comparison on the Basic Benchmark.} PlenopticDreamer generates high-fidelity visuals with consistent hallucinations from different camera trajectories. In contrast, ReCamMaster and TrajectoryCrafter fail to preserve spatio-temporal consistency while maintaining visual quality, especially under large-angle viewpoint changes, such as leftward azimuth shifts.}
\label{fig:main_comparison}
\vspace{-1em}
\end{figure*}

\noindent\textbf{Self-conditioned Training.}
When the total generation length $N$ becomes large, multiple inference steps are required, where previously generated videos are repeatedly used as conditioning inputs.
This recursive dependency can lead to error accumulation due to propagation of imperfect generations.
To alleviate this issue, we adopt a self-conditioned training strategy.
Specifically, in the first training stage of~\eqref{eq:loss}, all conditioning videos are ground-truth samples.
After convergence, the model is used to generate synthetic outputs from training-set input, which then replace the ground-truth conditions in the second training round.
This iterative refinement improves model robustness to imperfect inputs during long-range inference.

\section{Experiment}

\subsection{Experiment Setting}

\noindent\textbf{Implementation Details.}
We adopt Cosmos-Predict2.5-2B~\cite{ali2025world} as the backbone.
The generated videos have a resolution of 432$\times$768 with 93 frames.
We employ context parallelism~\cite{agarwal2025cosmos,ali2025world} to alleviate memory overhead and set the parallelism size to 8. Finetuning is conducted on 32 NVIDIA H100 GPUs with batch size 1 and a learning rate 2e-5. During finetuning, only the self-attention layers and camera encoder are updated, while all other parameters remain frozen. We post-train two model variants in~\secref{sec:basic_experiment} and~\secref{sec:agibot_experiment}.

\noindent\textbf{Evaluation Metrics.}
We evaluate models from three aspects:
1)~\textit{Visual Quality}: PSNR and FVD measure pixel- and frame-level fidelity, respectively.
2)~\textit{Camera Accuracy}: TransErr and RotErr~\cite{he2025cameractrl} quantify translation and rotation errors.
Dynamic poses are evaluated with ViPE~\cite{huang2025vipe}, while static novel views (\eg, azimuth/elevation shifts) accessed with VGGT~\cite{wang2025vggt} for relative pose estimation.
3)~\textit{Video Synchronization}: RoMa~\cite{edstedt2024roma} computes the number of matched pixels above a confidence threshold, denoted as Mat.
Pix.

\noindent\textbf{Baselines.}
We compare the proposed PlenopticDreamer with state-of-the-art camera-controlled generative video re-rendering methods: ReCamMaster~\cite{bai2025recammaster}, TrajectoryCrafter~\cite{yu2025trajectorycrafter}, and Trajectory-Attention~\cite{xiao2024trajectory}.
All baselines are used with their best-performing settings from the official open-sourced models.
For a fairer comparison, we also retrain ReCamMaster on Cosmos-Predict2.5 with Plücker raymaps on the same datasets, denoted as ReCamMaster*.

\subsection{Experiment on Basic Benchmark} \label{sec:basic_experiment}

\noindent\textbf{Experiment Details.}

\begin{itemize}
\item \underline{Functionality}: The model performs third-view to third-view transformations, such as left/right rotations, azimuth and elevation shifts, distance variations, and dynamic focal length changes.
\item \underline{Training Dataset}: We use MultiCamVideo~\cite{bai2025recammaster} and SynCamVideo~\cite{bai2024syncammaster}, large-scale synthetic datasets comprising approximately 136K and 34K episodes, respectively, depicting human motion captured under dynamic and static camera trajectories across 40 synthetic 3D environments.
\item \underline{Training Details}: The model context size $k$ is set to 4.
In the first stage, it is progressively trained for 10K, 4K, 1K, and 1K steps with context sizes 1–4, respectively.
In the second stage, the model generates synthetic data from 1,000 scenes and is further trained for 2K steps.
\item \underline{Benchmark}: We construct a Basic benchmark of 100 in-the-wild videos and 12 sequential camera trajectories.
\end{itemize}

\noindent\textbf{Qualitative and Quantitative Comparison.}
As shown in~\figref{fig:main_comparison} and~\tabref{tab:main_comparison}, PlenopticDreamer achieves superior view synchronization with high-fidelity visuals compared to all baselines (see the painting and electrical outlet on the wall in the first example, the traffic light in the second, and the eave above the robot in the third).
TrajectoryCrafter and Trajectory-Attention leverage 3D point tracking to extract dynamic cues from the source video and feed them as conditional inputs to the generator.
However, without updating the 3D memory using newly rendered content, they fail to maintain consistent cross-view synthesis.
Moreover, the off-the-shelf checkpoints of these baselines exhibit low camera accuracy, especially in translation, due to poor performance on static novel-view synthesis under large-angle viewpoint changes (\eg, azimuth and elevation shifts).
When retrained on the same datasets, ReCamMaster* achieves comparable camera accuracy.
Notably, the original ReCamMaster does not employ Plücker raymaps; we integrate them to ensure a fair comparison.

\subsection{Experiment on Agibot Benchmark} \label{sec:agibot_experiment}

\noindent\textbf{Experiment Details.}

\begin{itemize}
\item \underline{Functionality}: The model supports head-view to gripper-view transformations in robotic manipulation.
\item \underline{Training Dataset}: We use Agibot~\cite{bu2025agibot}, a large-scale robotic dataset with about 1M episodes. We sample 145,820 episodes, each containing three synchronized video views (one head-view and two gripper-views) with precise camera pose annotations.
\item \underline{Training Details}: The model context size $k$ is set to 2, and it is trained for 15K steps with merely the first stage, requiring $\sim$5 days.
\item \underline{Benchmark}: We build an Agibot benchmark using 200 test videos, covering head-to-hand and hand-to-hand camera transformations.
\end{itemize}

\begin{figure}[!t]
\vspace{-2em}
\centering
\includegraphics[width=0.50\textwidth]{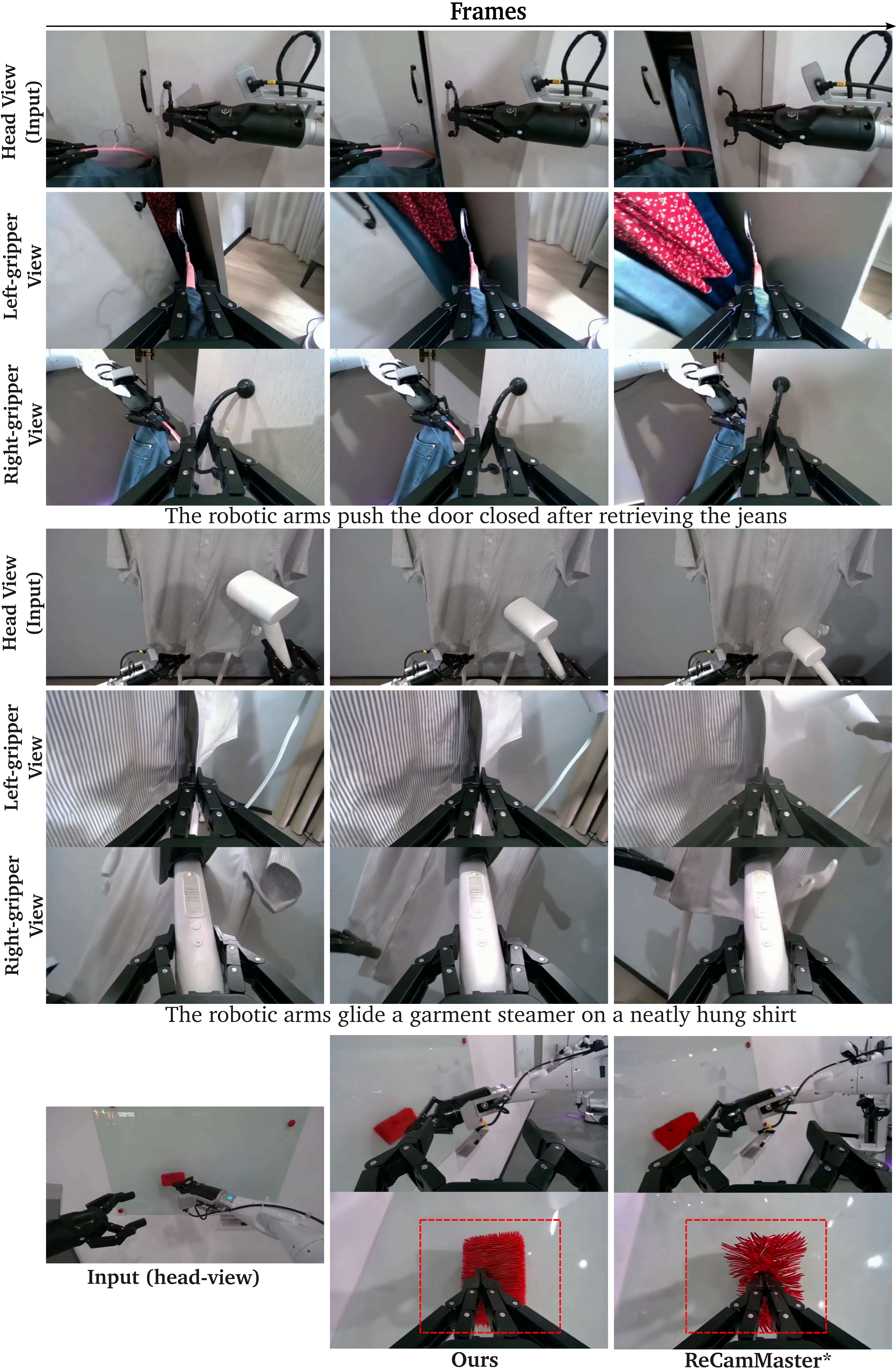}
\caption{\textbf{Qualitative Results on the Agibot Benchmark.} Given a head-view manipulation video in Agibot, PlenopticDreamer-agibot (Ours) can generate temporally consistent videos from the left and right gripper viewpoints.}
\label{fig:agibot}
\vspace{-2em}
\end{figure}

\noindent\textbf{Qualitative and Quantitative Comparison.}
As illustrated in~\figref{fig:agibot} and~\tabref{tab:agibot}, PlenopticDreamer can perform head-view→gripper-view and gripper-view→gripper-view transformation in an autoregressive manner.
Specifically, given a head-view manipulation video, it generates temporally consistent videos from both left and right gripper viewpoints across diverse manipulation tasks.
In contrast, ReCamMaster* (also retrained on the Agibot dataset) fails to maintain view synchronization and high visual quality (see the blackboard eraser marked in the red dashed box).

\begin{table}[bh!] 
\caption{\textbf{Quantitative Comparison on the Agibot Benchmark}. Visual quality and view synchronization are assessed on 2 shots (left and right gripper viewpoints).} \label{tab:agibot} 
\centering 
\begin{threeparttable}
\resizebox{.45\textwidth}{!} 
{ \begin{tabular}{ccccccccc} 
\toprule 
&  PSNR $\uparrow$ & View Sync. (Mat. Pix.(K) $\uparrow$) \\
\midrule

ReCamMaster* & 13.84 & 13.2 \\ 
Ours & \textbf{14.54} & \textbf{15.3} \\
\bottomrule 
\end{tabular} } 
\end{threeparttable} 
\vspace{-2em}
\end{table}

\begin{figure}[!tp]
\centering
\includegraphics[width=0.50\textwidth]{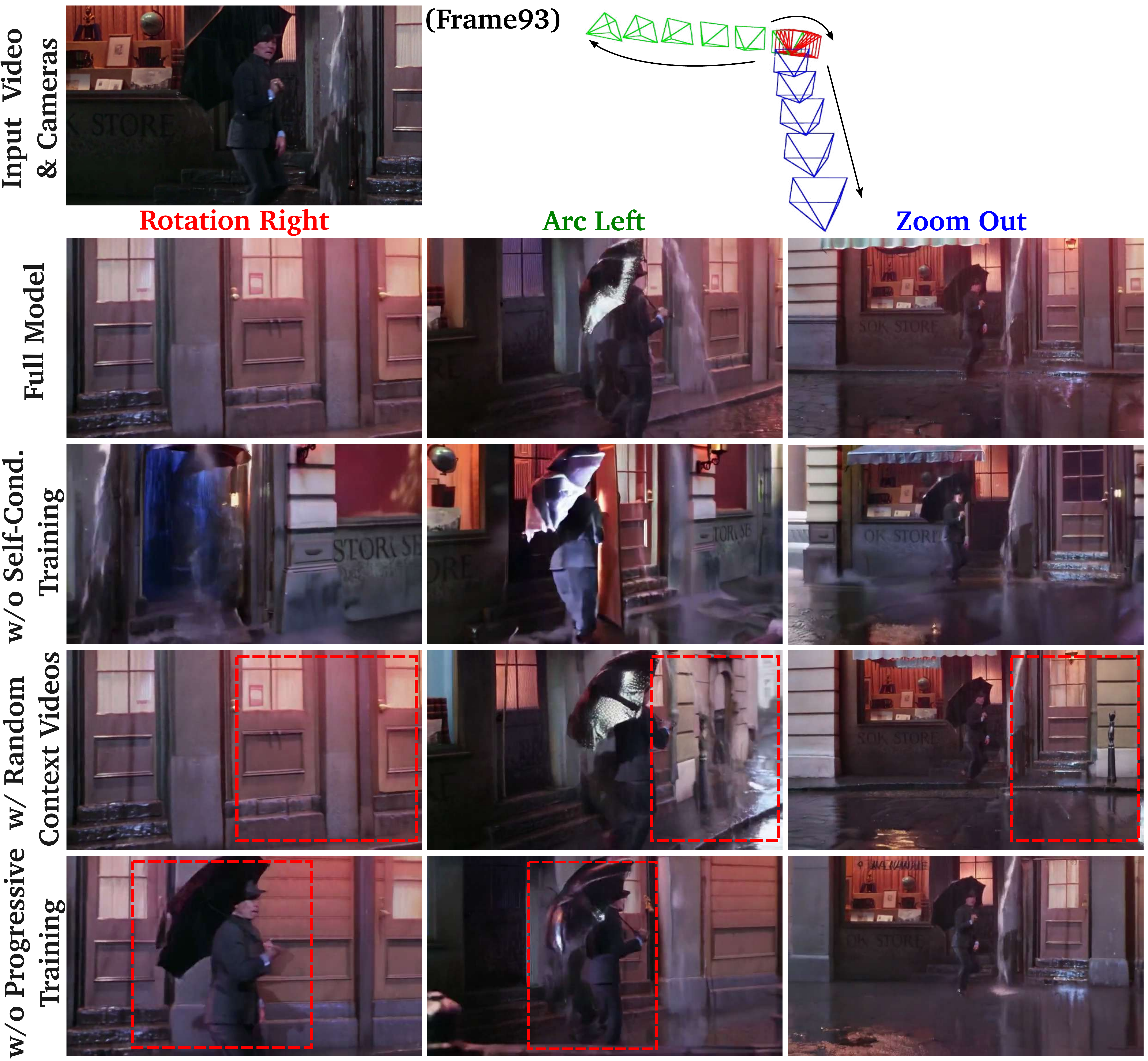}
\caption{\textbf{Ablation Study.} Qualitative visualization of effects from different training strategies and context retrieval method.}
\label{fig:ablation_full}
\vspace{-2em}
\end{figure}

\begin{table*}[ht!] 
\caption{\textbf{Ablation Study on the Basic Benchmark.} Evaluation is conducted on the full set, comprising a total of 1,200 generated videos.}
\label{tab:main_ablation} 
\centering 
\begin{threeparttable}
\resizebox{.95\textwidth}{!} 
{ \begin{tabular}{ccccccccc} 
\toprule 
& \multicolumn{2}{c}{ Visual Quality } & \multicolumn{2}{c}{ Camera Accuracy } & \multicolumn{4}{c}{ View Synchronization (Mat. Pix.(K) $\uparrow$)} \\ \cmidrule(lr){2-3} \cmidrule(lr){4-5} \cmidrule(lr){6-9} Model & FVD $\downarrow$ & IQ$\uparrow$ & TransErr $\downarrow$ & RotErr (rad) $\downarrow$ & 3 Shots & 6 Shots & 9 Shots & 12 Shots \\ 
\midrule 
w/o Self-Cond. Training & 464.3 & 56.7 & \textbf{0.54} & 0.23 & \underline{40.9} & 40.2 & \underline{45.1} & \underline{40.7} \\
w/ Random Context Retrieval & 520.5 & \underline{58.3} & \underline{0.56} & \textbf{0.20} & 33.6 & 33.4 & 36.5 & 32.4 \\ 
w/o Progressive Training & \underline{453.8} & 57.2 & 0.63 & 0.23 & 39.6 & \underline{40.6} & 43.6 & 39.4 \\ 
\midrule
Full Model & \textbf{425.8} & \textbf{58.5} & \textbf{0.54} & \underline{0.21} & \textbf{41.4} & \textbf{40.8} & \textbf{45.4} & \textbf{41.2} \\ 
\bottomrule 
\end{tabular} } 
\end{threeparttable} 
\label{tab:ablation_full}
\vspace{-1em}
\end{table*}
\begin{table}[htb!] 
\caption{\textbf{Ablation on Retrieved Context Video Number}.} \label{tab:context_view_num} 
\centering 
\begin{threeparttable}
\resizebox{.42\textwidth}{!} 
{ \begin{tabular}{ccccccccc} 
\toprule 
& \multicolumn{4}{c}{View Synchronization (Mat. Pix.(K) $\uparrow$)} \\ \cmidrule(lr){2-5} 
Video Num. & 3 Shots & 6 Shots & 9 Shots & 12 Shots \\ 
\midrule

4 & \textbf{58.1} & 51.2 & 52.1 & 42.7 \\ 
6 & \multirow{3}{*}{
\tikz{\draw[->, line width=0.6pt] (0,0) -- (0,1.2);}
} & \textbf{52.1} & \textbf{53.1} & \textbf{43.6} \\ 
8 &  & \multirow{2}{*}{
\tikz{\draw[->, line width=0.6pt] (0,0) -- (0,0.8);}
} & 50.2 & 41.0 \\ 
10 &  & & \multirow{1}{*}{
\tikz{\draw[->, line width=0.6pt] (0,0) -- (0,0.4);}} & 40.8 \\ 

\bottomrule 
\end{tabular} } 
\end{threeparttable} 
\vspace{-2.2em}
\end{table}

\begin{figure*}[!h]
\centering
\includegraphics[width=0.99\textwidth]{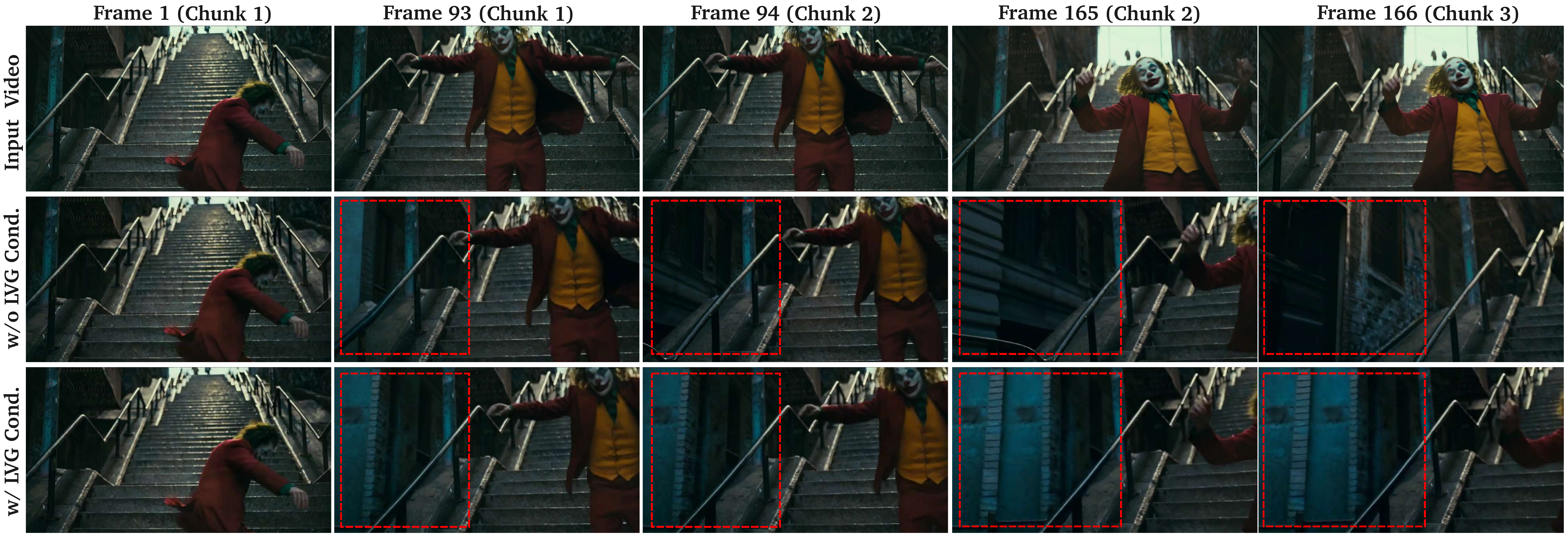}
\caption{\textbf{Long Video Generation.} Given a leftward rotation camera trajectory, ours (w/ LVG Cond.) preserves spatial consistency across adjacent video chunks, yielding seamless transitions at their boundaries (highlighted by red dotted lines).}
\label{fig:lvg}
\vspace{-1em}
\end{figure*}
\begin{figure}[!tp]
\centering
\includegraphics[width=0.50\textwidth]{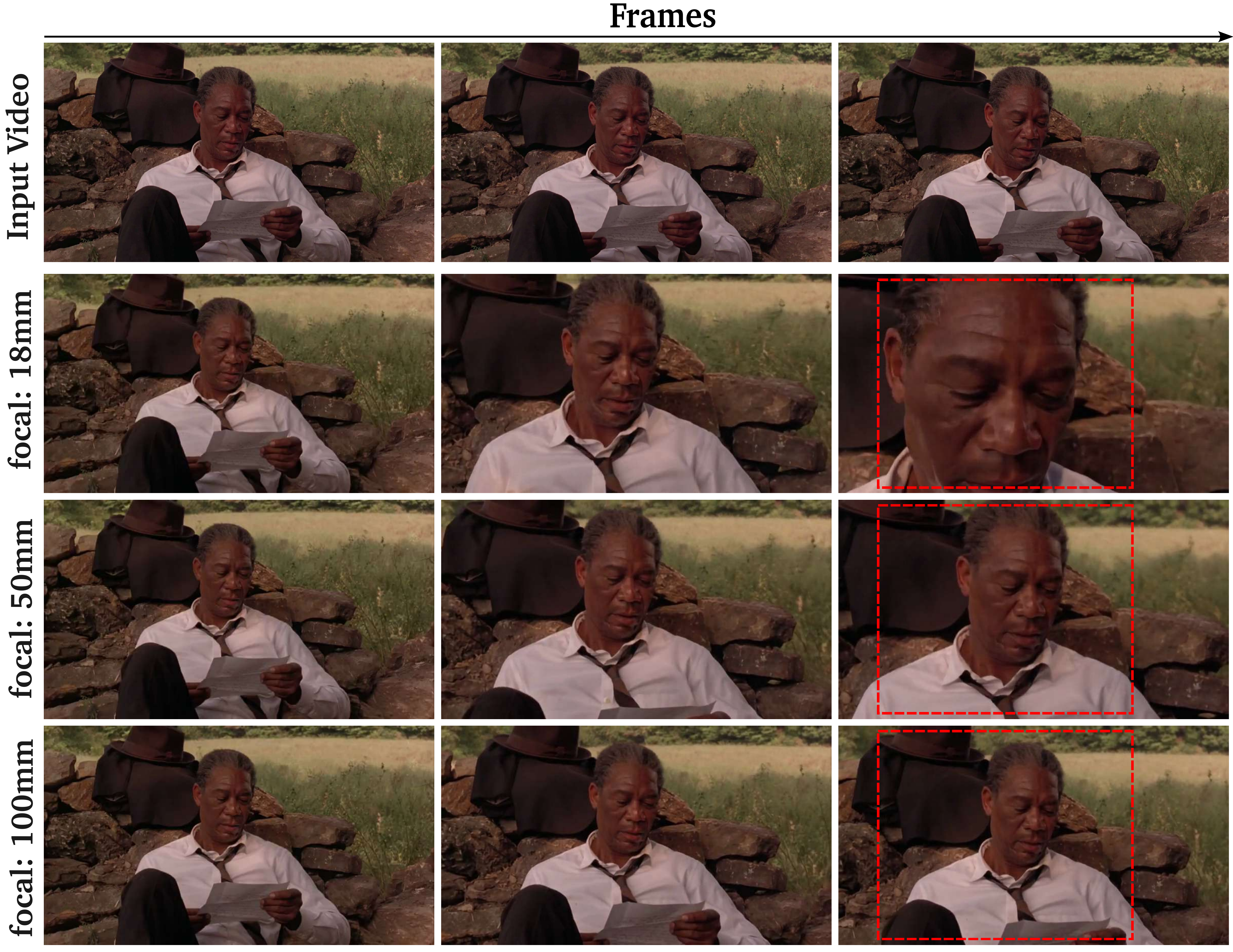}
\caption{\textbf{Focal Length Effect.} Our method simulates varying depth-of-field effects corresponding to different focal lengths (18mm$\rightarrow$100mm) under a “zoom-in” camera trajectory.}
\label{fig:focal_length}
\vspace{-2em}
\end{figure}

\subsection{Ablation Study}

\noindent\textbf{Training Strategy.}
As shown in~\figref{fig:ablation_full} and~\tabref{tab:ablation_full}, removing the progressive training strategy (\textit{w/o Progressive Training}) leads to a notable degradation in camera accuracy (0.54$\rightarrow$0.63 in TransErr), and an occluded man becomes erroneously visible under the “Rotation Right” case.
This indicates that progressively enlarging the context size effectively stabilizes model convergence and enhances camera performance.
When the self-conditioned training is removed, the generated videos exhibit pronounced artifacts and over-exposure, particularly in long-shot sequences.
Correspondingly, both FVD and IQ (Image Quality in VBench~\cite{huang2024vbench}) metrics worsen, verifying that training with imperfect inputs enhances model robustness and mitigates error accumulation over time.

\noindent\textbf{Video Retrieval Strategy.}
Replacing the proposed retrieval mechanism with random selection leads to a significant decline in view synchronization across all shots, with inconsistent hallucinations (highlighted with red dashed boxes in~\figref{fig:ablation_full}).
We further examine the impact of context size in~\tabref{tab:context_view_num}: increasing the number of retrieved contexts from 4 to 6 enhances multi-view consistency by offering richer spatial cues.
However, further enlarging the context brings diminishing gains due to compounded trajectory fusion errors and accumulated generative noise.

\subsection{Application}

\noindent\textbf{Long Video Generation.}
With the proposed long-video conditioning strategy, PlenopticDreamer supports coherent long-context video re-rendering, as shown in~\figref{fig:lvg}.
Given a leftward rotation trajectory, our method produces temporally consistent long video segments while preserving spatial coherence across adjacent chunks.
In contrast, removing this conditioning results in visible inconsistencies, as illustrated in the second row.

\noindent\textbf{Focal Length Effect.}
Varying the input focal length leads to corresponding depth-of-field changes, as shown in~\figref{fig:focal_length} under a “zoom-in” camera trajectory.
This enables finer control over camera behavior and visual emphasis, offering users greater flexibility in camera-aware video generation.

\section{Conclusion}

We introduce PlenopticDreamer, a camera-controlled generative video re-rendering framework enforcing spatio-temporal consistency.
It employs a multi-in-single-out,~\textit{autoregressive} diffusion model conditioned on spatio-temporal memory, retrieved via 3D FOV strategy for coherent hallucinations along trajectories.
By incorporating progressive context-scaling and self-conditioning during training, the method improves stability and reduces error accumulation in long-range video generation.
Evaluation on Basic and Agibot benchmarks shows state-of-the-art view synchronization, high fidelity, and precise camera control.

\noindent\textbf{Limitations.}
Despite self-conditioned training, ours still exhibits occasional failures, including over-exposure and distortion in long-shot videos.
Future work could explore a Self-Forcing–style~\cite{huang2025self,liu2025rolling,cui2025self} paradigm.
We also observe degraded performance in complex human motions, such as dancing, likely from pretraining data biases in Cosmos.

\setcounter{table}{0}
\setcounter{figure}{0}
\renewcommand{\thetable}{R\arabic{table}}
\renewcommand\thefigure{S\arabic{figure}}

\appendix

\section{More Experimental Details}

\subsection{Implementation} 

\noindent \textbf{Video Retrieval Algorithm.}
To construct the view frustum for a single camera pose, we fix the horizontal and vertical fields of view to 90$^\circ$ and 60$^\circ$, respectively, and set the near and far clipping planes to 0 and 10. For mesh sampling, we uniformly sample 8 points along the width and 6 points along the height on the plane.

\noindent \textbf{Choice of Context Number $k$.}
A larger value for $k$ allows for the retrieval of more contextual information and reduces the number of required inference iterations, but it also increases computational overhead.  but comes at the cost of increased computational overhead. To balance context visibility and computation, we adopt video consistency as the selection criterion, as shown in~\tabref{tab:context_k_num}, and choose $k=4$ as the appropriate setting.

\begin{table}[htb!] 
\caption{\textbf{Ablation on In-context Video Number $k$}. View Synchronization is evaluated on the Basic Benchmark.} \label{tab:context_k_num} 
\centering 
\begin{threeparttable}
\resizebox{.43\textwidth}{!} 
{ \begin{tabular}{ccccccccc} 
\toprule 
N Shots/$k$ videos & 2 & 3 & 4 & 5 & 6 \\ 
\midrule
6 & 38.9 & 39.7 & \underline{40.3} & \textbf{40.8} & 40.2\\
9 & 43.6 & 44.5 & \textbf{45.6} & \underline{45.4} & 44.7\\
12 & 40.8 & 40.2 & \textbf{41.2} & \underline{40.9} & 40.4\\
\bottomrule 
\end{tabular} } 
\end{threeparttable} 
\end{table}

\noindent \textbf{Long Video Generation.}
During training, we adopt 6 overlapped latent frames (corresponding to 21 decoded frames) and set the conditioning ratio to 0.45. Although the model is trained on 81-frame multi-view datasets, it generalizes effectively to long-form video generation. During inference, we produce a 93-frame initial chunk, followed by subsequent chunks of 71 frames.

\noindent \textbf{Self-Conditioned Training.}
For the second training stage of the model evaluated on the Basic benchmark, we randomly sample 900 scenes from the MultiCamVideo dataset and 100 scenes from the SynCamVideo dataset. For each scene, we synthesize 1–5 videos, yielding roughly 3.5K training samples in total. In our current setup, the clean ground-truth video is used as the context for generating its noisy pseudo–GT counterpart. We did not observe clear performance gains when incorporating long-shot autoregressive generation into the synthetic video pipeline, as reported in~\tabref{tab:synthetic_n_shots}. For each $N$-shot setting, we generate the same number of synthetic videos and post-train the model for 2K steps to ensure a fair comparison.

\begin{table}[htb!] 
\caption{\textbf{Ablation on $N$-Shots Synthetic Video Generation}. Evaluation is conducted on the Basic Benchmark.} \label{tab:synthetic_n_shots} 
\centering 
\begin{threeparttable}
\resizebox{.38\textwidth}{!} 
{ \begin{tabular}{ccccccccc} 
\toprule 
 & 1 Shot & 2 Shots & 3 Shots & 4 Shots \\ 
\midrule
FVD$\downarrow$ & \textbf{425.8} & 441.3 & \underline{436.4} & 460.2\\
IQ$\uparrow$ & \textbf{58.5} & \underline{57.6} & 57.2 & 56.5\\
\bottomrule 
\end{tabular} } 
\end{threeparttable} 
\vspace{-1em}
\end{table}

\subsection{Evaluation}

\noindent \textbf{FVD (Fréchet Video Distance).}
We use StyleGAN‑V~\cite{skorokhodov2022stylegan} as the feature extractor backbone, sample 49 frames per video interval, resize each frame to $432 \times 768$, and include all frames for evaluation. For videos with substantial camera motion (low FOV overlap) relative to the input video, we select alternative video with similar camera trajectories for evaluation. Thus this metric can provide an indirect measure of video similarity.

\noindent \textbf{TransErr (Camera Translation Error).}
\begin{equation}
\text { TransErr }=\sum_{i=1}^n\left\|\mathbf{T}_{g t}^i-\mathbf{T}_{pred}^i\right\|_2^2
\end{equation}
where $\mathbf{T}_{gt}^i$ and $\mathbf{T}_{pred}^i$ are the ground-truth and predicted translation vectors for the $i$-th frame. In our experiments, we first align the translation scale of cameras estimated by ViPE~\cite{huang2025vipe} or VGGT~\cite{wang2025vggt} with the input cameras before computing TransErr.

\noindent \textbf{RotErr (Camera Rotation Error).}
\begin{equation}
\text { RotErr }=\sum_{i=1}^n \arccos \frac{\left.\operatorname{tr}\left(\mathbf{R}_{gt}^i \mathbf{R}_{pred}^{i T}\right)\right)-1}{2}
\end{equation}
where $\mathbf{R}_{gt}^i$ and $\mathbf{R}_{pred}^i$ are the ground-truth and predicted rotation matrices for the $i$-th frame, and $\operatorname{tr}(\cdot)$ denotes the matrix trace. We report this metric in radians.

\begin{figure}[!tp]
\centering
\includegraphics[width=0.46\textwidth]{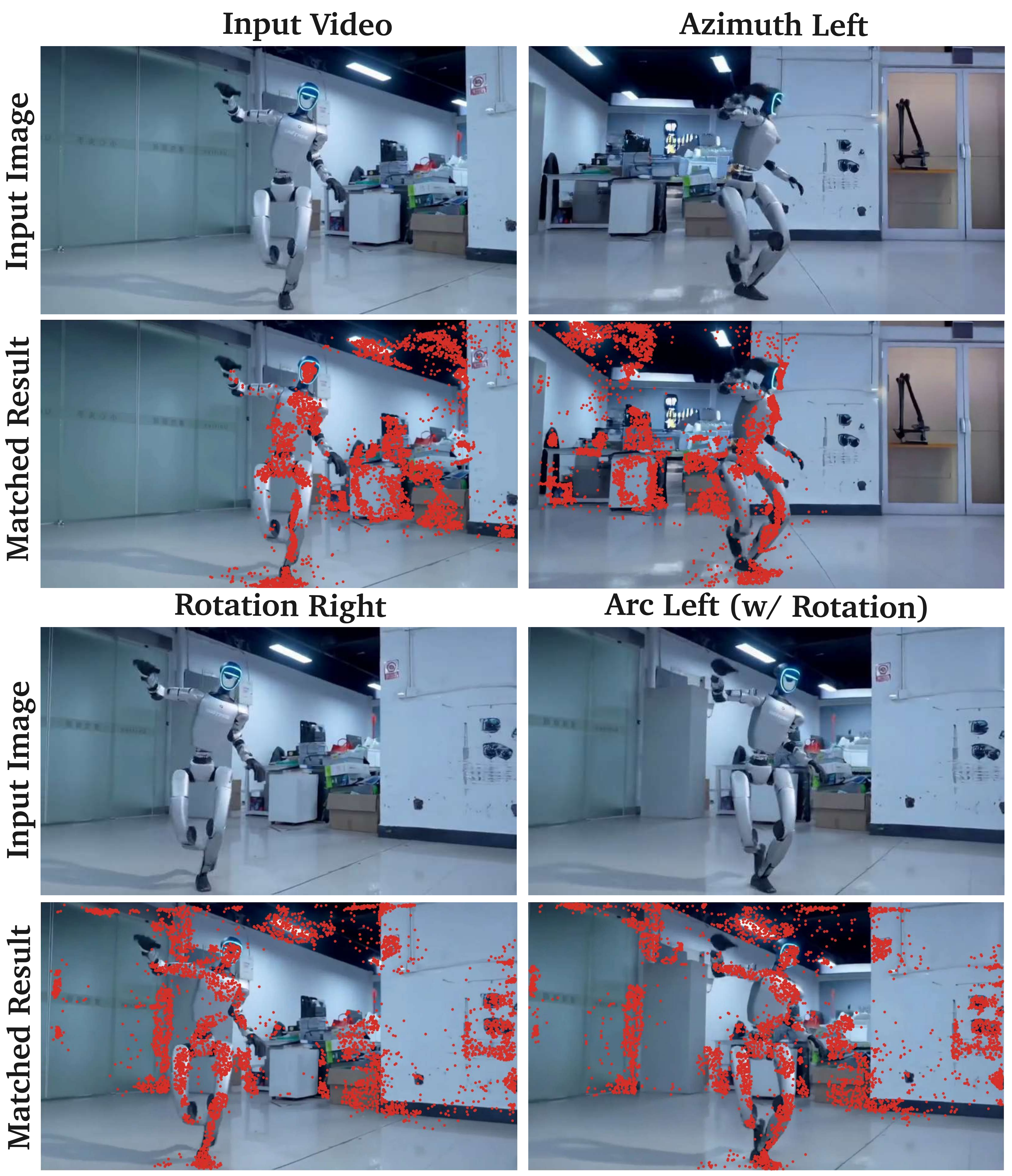}
\caption{\textbf{Image Matching Result.} The red points indicate the matched pixel correspondences across the input images..}
\label{fig:img_match}
\vspace{-1.4em}
\end{figure}

\noindent \textbf{Mat. Pix. (Matched Pixels in Video Synchronization).}
\begin{equation}
\text{Mat. Pix.} = \sum_{i=1}^{K} \mathbf{1}\big( C_i \ge \tau \big)
\end{equation}
where $K$ is the total number of pixels, $C_i$ is the confidence score of the $i$-th pixel, $\tau$ is the confidence threshold, and $\mathbf{1}(\cdot)$ is the indicator function. Mat. Pix. counts pixels with confidence above the threshold. The qualitative matching results are presented in~\figref{fig:img_match}.

\begin{figure*}[!th]
\centering
\includegraphics[width=.94\textwidth]{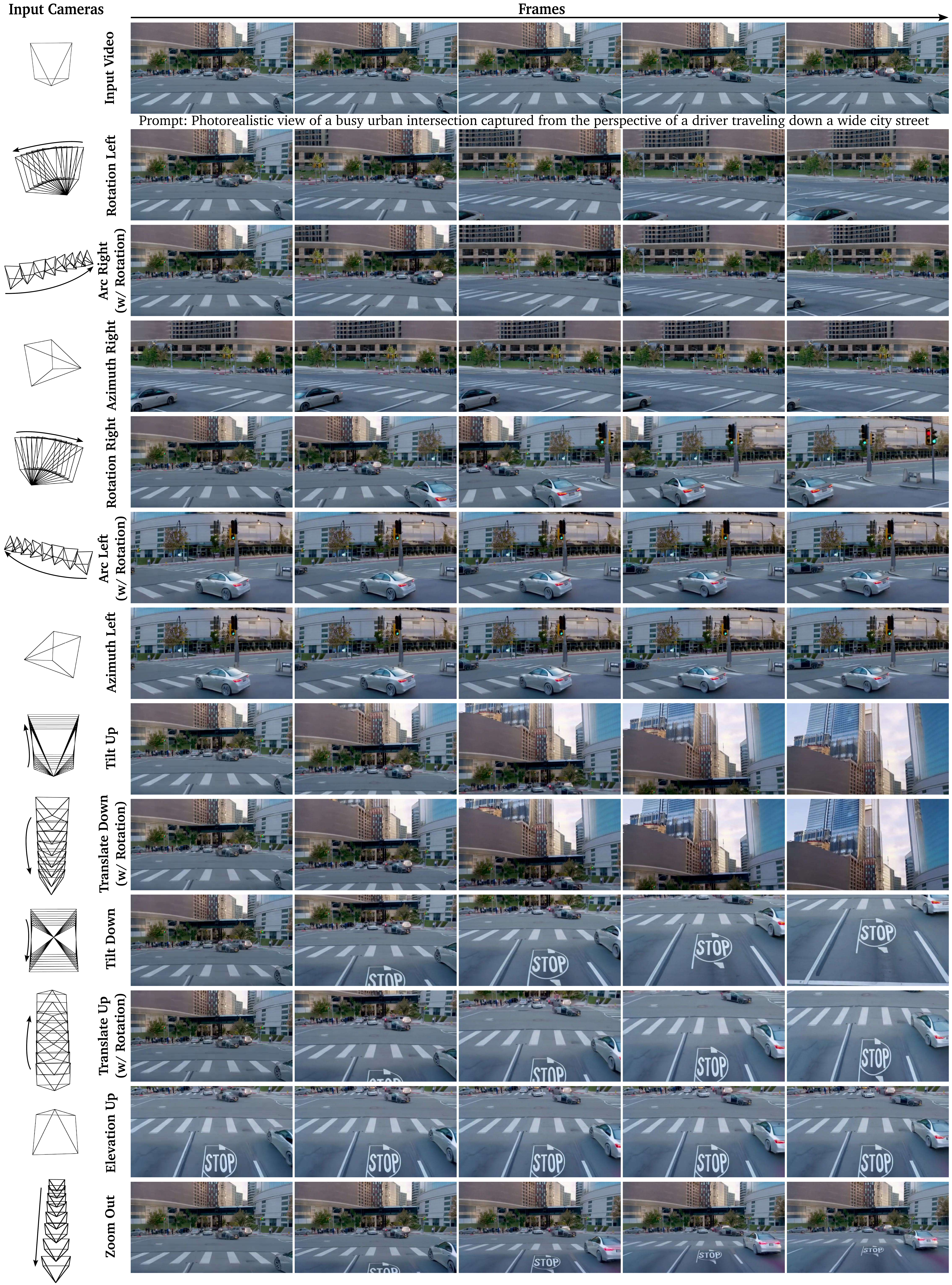}
\vspace{-1.2em}
\caption{\textbf{Full Camera Trajectories on the Basic Benchmark.} The sequence proceeds as: (1) Rotation Left$\rightarrow$(2) Arc Right (w/ Rot.)$\rightarrow$(3) Azimuth Right$\rightarrow$(4) Rotation Right$\rightarrow$(5) Arc Left (w/ Rot.)$\rightarrow$(6) Azimuth Left$\rightarrow$(7) Tilt Up$\rightarrow$(8) Translate Down (w/ Rot.)$\rightarrow$(9) Tilt Down$\rightarrow$(10) Translate Up (w/ Rot.)$\rightarrow$(11) Elevation Up$\rightarrow$(12) Zoom Out.}
\label{fig:full_basic_cameras}
\end{figure*}

In our experiments, we set $\tau = 0.5$, resize frames to $432 \times 768$, and average all frames. As illustrated in~\figref{fig:full_basic_cameras}, the sequential 12-shot trajectory is:  
(1) Rotation Left $\rightarrow$ (2) Arc Right (w/ Rot.) $\rightarrow$ (3) Azimuth Right $\rightarrow$ (4) Rotation Right $\rightarrow$ (5) Arc Left (w/ Rot.) $\rightarrow$ (6) Azimuth Left $\rightarrow$ (7) Tilt Up $\rightarrow$ (8) Translate Down (w/ Rot.) $\rightarrow$ (9) Tilt Down $\rightarrow$ (10) Translate Up (w/ Rot.) $\rightarrow$ (11) Elevation Up $\rightarrow$ (12) Zoom Out. View synchronization is computed for the video pairs shown in~\tabref{tab:video_sync_pairs}.

\begin{table}[bh!] 
\caption{Video Pairs for Multi-shot Video Synchronization Calculation on the Basic Benchmark.} 
\label{tab:video_sync_pairs} 
\centering 
\begin{threeparttable}
\resizebox{.4\textwidth}{!} 
{ 
\begin{tabular}{cc} 
\toprule 
N-Shots & Calculated Video Pairs \\
\midrule 
\multirow{2}{*}{3 Shots} & (Rotation Left, Arc Right (w/ Rot.))  \\
                         & (Rotation Left, Azimuth Right)  \\
\midrule
\multirow{4}{*}{6 Shots} & (Rotation Left, Arc Right (w/ Rot.))  \\
                         & (Rotation Left, Azimuth Right)  \\
                         & (Rotation Right, Arc Left (w/ Rot.))  \\
                         & (Rotation Right, Azimuth Left)  \\
\midrule
\multirow{6}{*}{9 Shots} & (Rotation Left, Arc Right (w/ Rot.))  \\
                         & (Rotation Left, Azimuth Right)  \\
                         & (Rotation Right, Arc Left (w/ Rot.))  \\
                         & (Rotation Right, Azimuth Left)  \\
                         & (Tilt Up, Translate Down (w/ Rot.))  \\
                         & (Tilt Down, Translate Up (w/ Rot.))  \\
\midrule
\multirow{8}{*}{12 Shots} & (Rotation Left, Arc Right (w/ Rot.))  \\
                         & (Rotation Left, Azimuth Right)  \\
                         & (Rotation Right, Arc Left (w/ Rot.))  \\
                         & (Rotation Right, Azimuth Left)  \\
                         & (Tilt Up, Translate Down (w/ Rot.))  \\
                         & (Tilt Down, Translate Up (w/ Rot.))  \\    
                         & (Translate Up (w/ Rot.), Elevation Up)  \\
                         & (Translate Up (w/ Rot.), Zoom Out)  \\ 
\bottomrule 
\end{tabular} 
} 
\end{threeparttable} 
\end{table}

\section{More Qualitative Results}
\begin{figure*}[!th]
\centering
\includegraphics[width=0.86\textwidth]{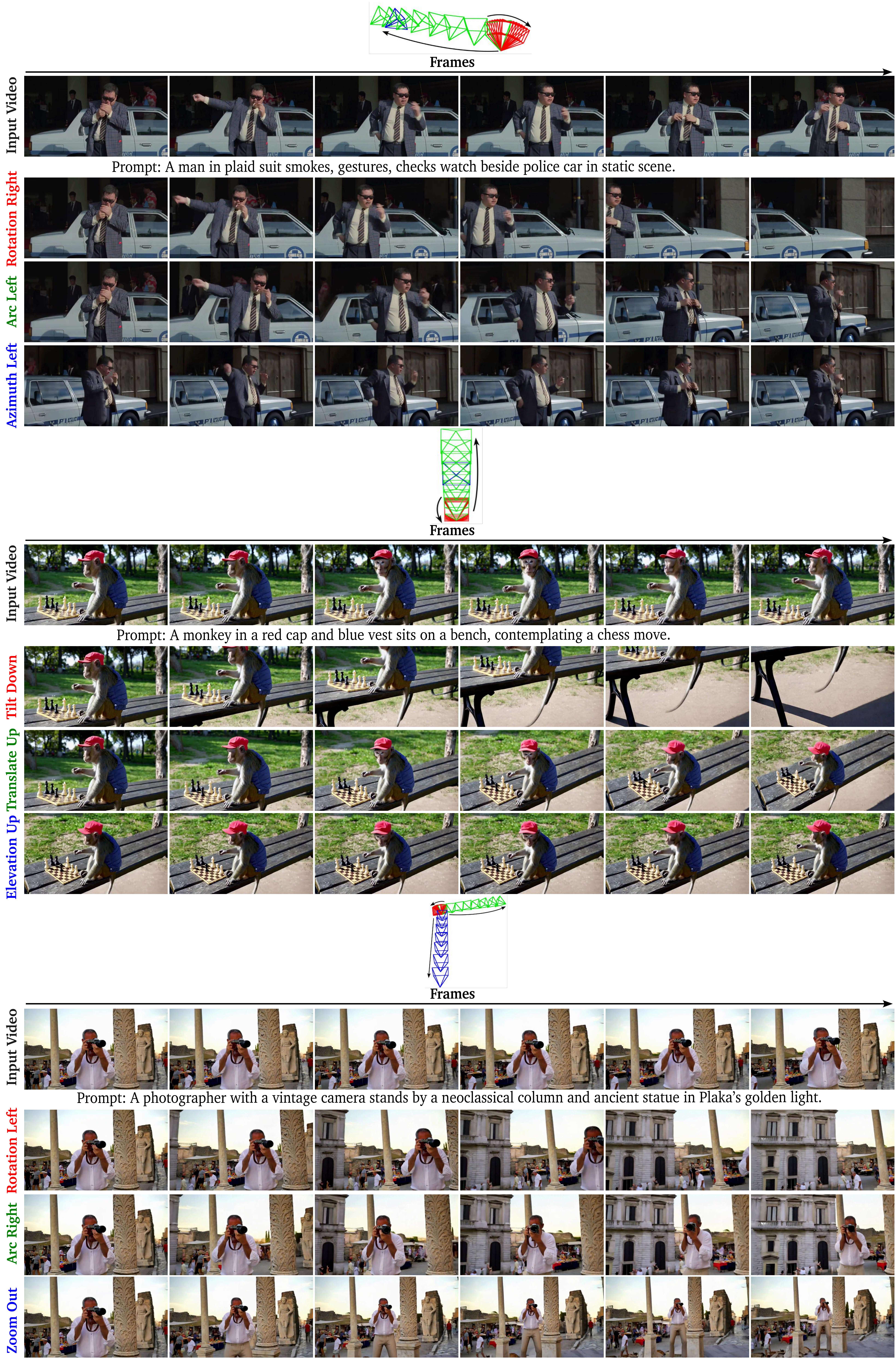}
\vspace{-0.8em}
\caption{\textbf{More Visual Results on the Basic Benchmark.} Our method generates consistent hallucinated context in unseen region.}
\label{fig:more_basic_result}
\end{figure*}

\begin{figure*}[!th]
\centering
\includegraphics[width=\textwidth]{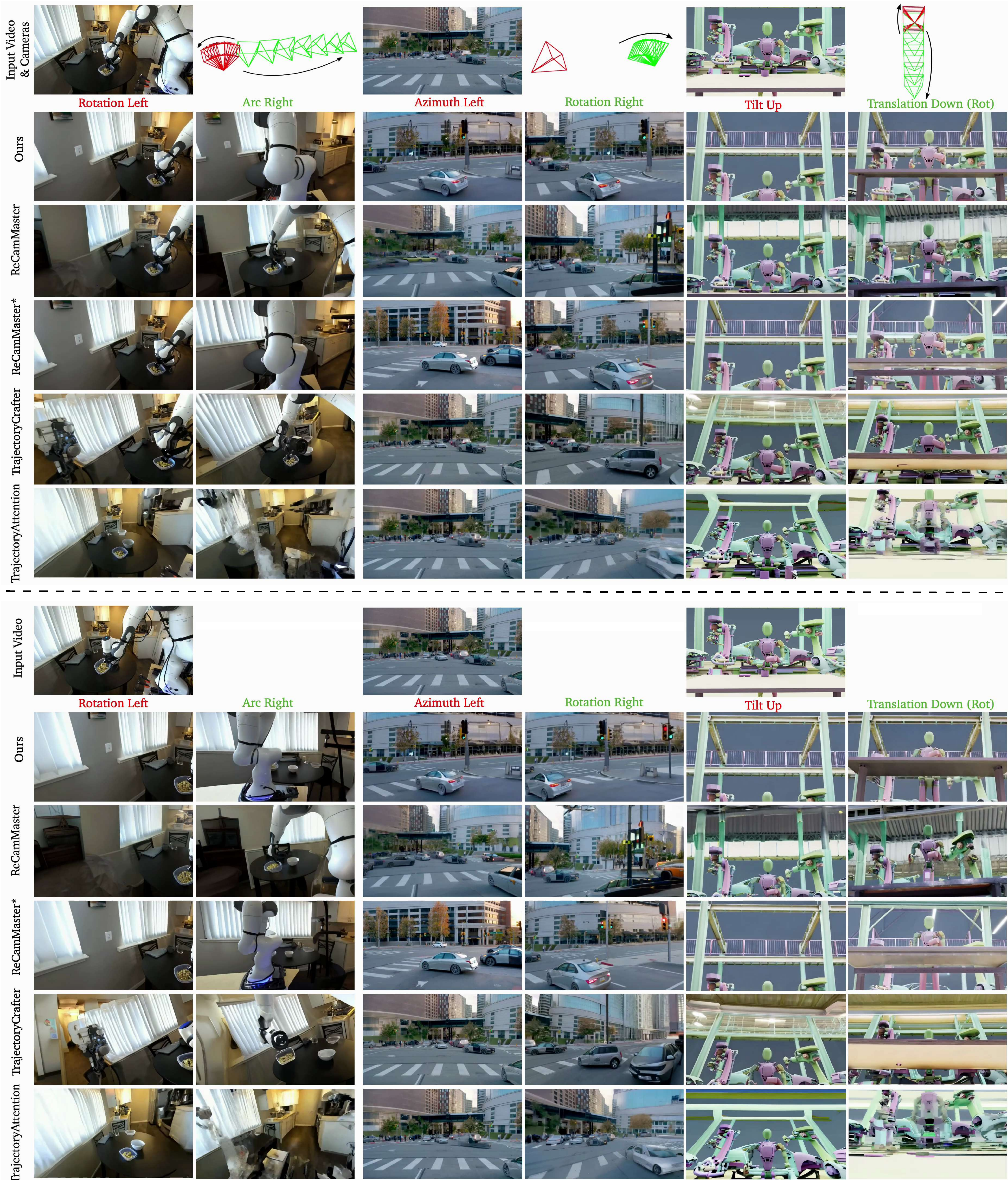}
\caption{\textbf{More Qualitative Comparison on the Basic Benchmark.} The figures above and below correspond to frames 54 and 88, respectively. Please check full videos on the website provided in the website.}
\label{fig:main_comparison_basic_supp}
\end{figure*}

\begin{figure*}[!th]
\centering
\includegraphics[width=0.91\textwidth]{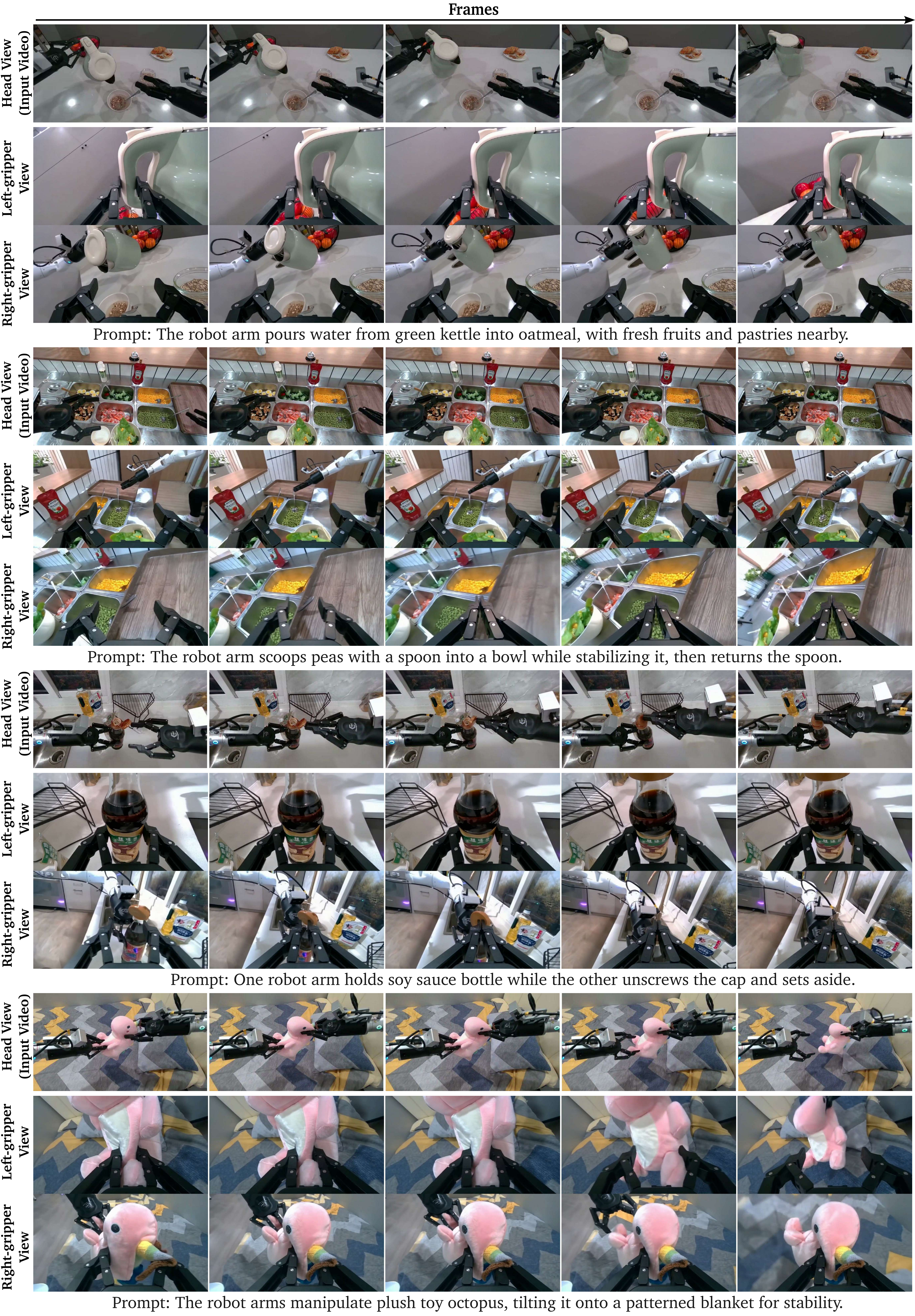}
\vspace{-0.8em}
\caption{\textbf{More Visual Results on the Agibot Benchmark.} The sequence proceeds as: (1) Left-gripper View$\rightarrow$(2) Right-gripper View.}
\label{fig:more_agibot_result}
\end{figure*}

\begin{figure*}[!th]
\centering
\includegraphics[width=.75\textwidth]{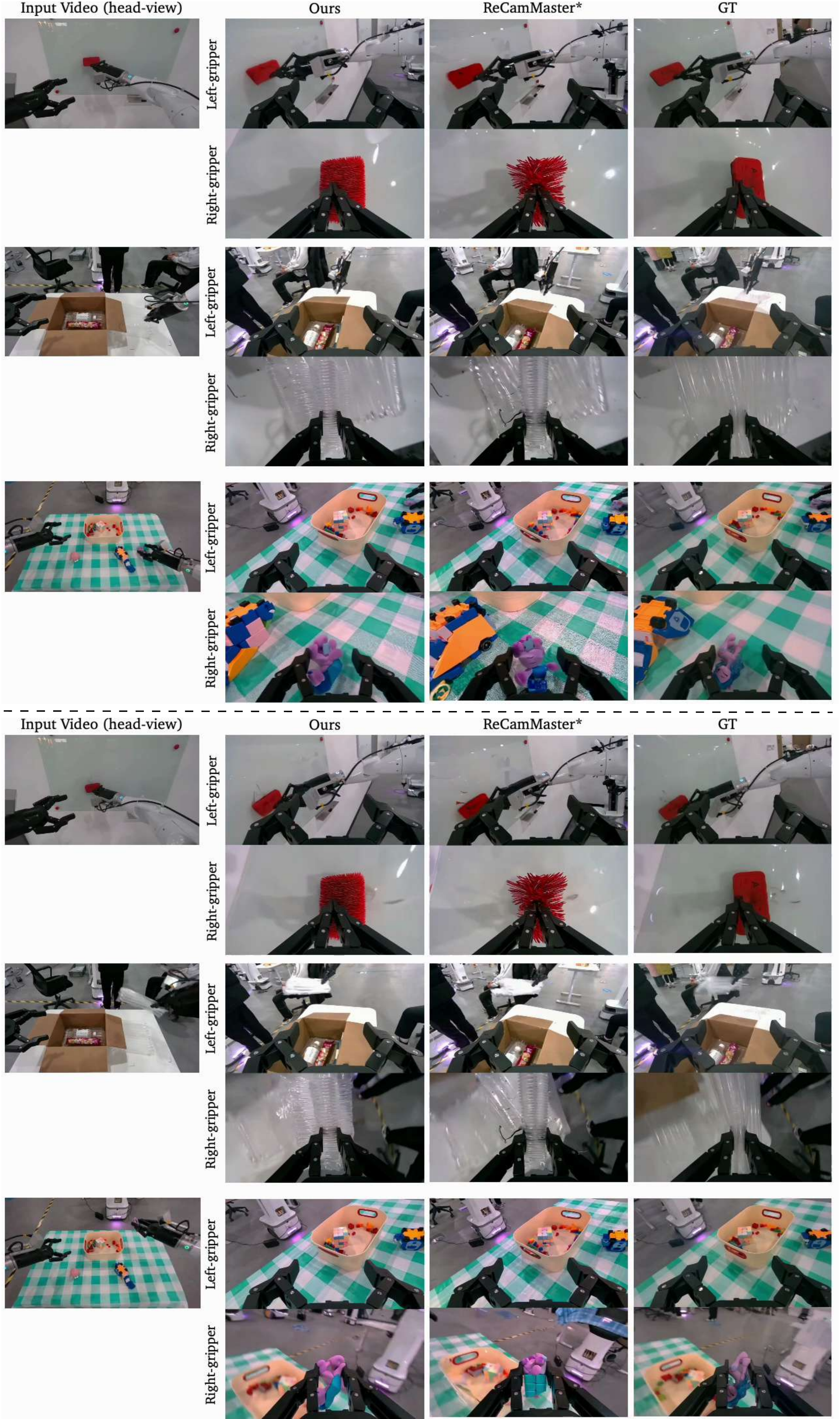}
\caption{\textbf{More Qualitative Comparison on the Agibot Benchmark.} The figures above and below correspond to frames 24 and 93, respectively. Compared to our method, ReCamMaster* exhibits noticeably stronger object distortion and inconsistency.}
\label{fig:main_comparison_agibot_supp}
\end{figure*}

\begin{figure*}[!th]
\centering
\includegraphics[width=.86\textwidth]{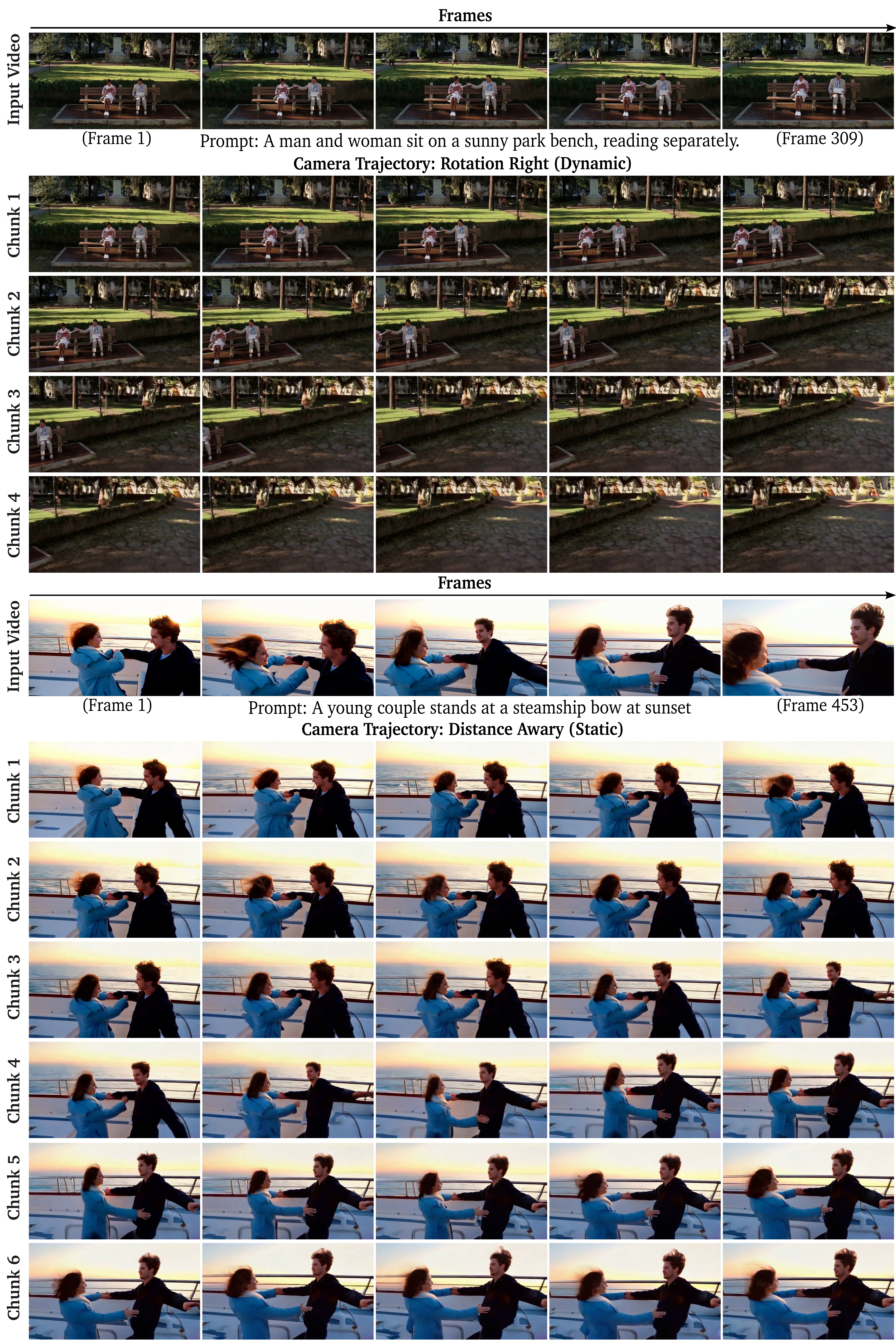}
\caption{\textbf{More Long Video Generation Results under Dynamic and Static Novel-Camera Settings.}}
\label{fig:lvg_supp}
\end{figure*}

\begin{figure*}[!th]
\centering
\includegraphics[width=.9\textwidth]{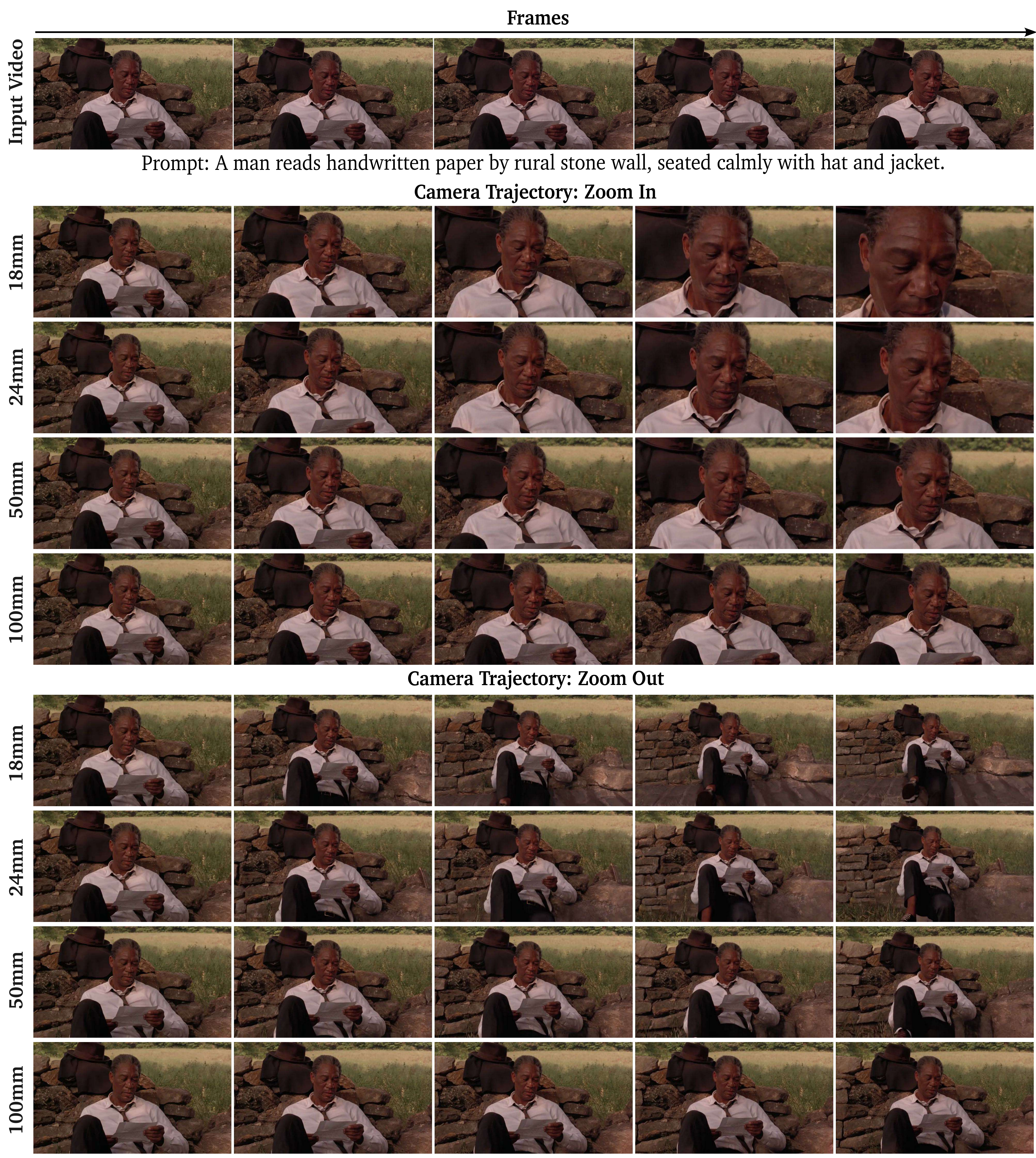}
\caption{\textbf{More Focal Length Effect Results.} Our method synthesizes depth-of-field variations across focal lengths (18mm$\rightarrow$100mm). Shorter focal lengths produce more greater changes in the resulting field-of-view (FOV).}
\label{fig:focal_length_supp}
\end{figure*}

We present additional qualitative results on the Basic Benchmark in~\figref{fig:more_basic_result}, along with comparative visualizations in~\figref{fig:main_comparison_basic_supp}. Further results on the Agibot Benchmark are shown in~\figref{fig:more_agibot_result}, with corresponding comparisons in~\figref{fig:main_comparison_agibot_supp}. We also include extended long video generation results in~\figref{fig:lvg_supp} and illustrate the focal-length effect in~\figref{fig:focal_length_supp}.


{
    \small
    \bibliographystyle{ieeenat_fullname}
    \bibliography{main}
}


\end{document}